\documentclass[journal=jacsat,manuscript=article]{achemso}

\usepackage[version=3]{mhchem} 

\author{Mrityunjay Sharma}
\altaffiliation{These authors contributed equally to this work.}
\affiliation[CSIR]
{CSIR- Central Scientific Instruments Organisation, Sector 30-C, Chandigarh-160030, India}
\alsoaffiliation[ACSIR]
{Academy of Scientific and Innovative Research (AcSIR), Ghaziabad-201002, India}
\alsoaffiliation[Prof]
{Department of Higher Education,Himachal Pradesh,Shimla -171001,India}
\email{mrityunjay.csio24j@acsir.res.in}
\author{Sarabeshwar Balaji}
\altaffiliation{These authors contributed equally to this work.}
\affiliation[IISERB]
{Indian Institute of Science Education and Research Bhopal(IISERB),
Madhya Pradesh-462066,India}
\author{Valentina Parma}
\affiliation[UM]
{Monell Chemical Senses Center, Philadelphia, Pennsylvania, USA 19104}
\alsoaffiliation[UMM]
{Department of Otolaryngology, Head and Neck Surgery, University of Pennsylvania, Philadelphia, Pennsylvania, USA 19104}

\author{Ritesh Kumar}
\email{riteshkr.csio@csir.res.in}
\affiliation[CSIR]
{CSIR- Central Scientific Instruments Organisation, Sector 30-C, Chandigarh-160030, India}
\alsoaffiliation[ACSIR]
{Academy of Scientific and Innovative Research (AcSIR), Ghaziabad-201002, India}

\abbreviations{IR,NMR,UV}
\usepackage{multirow}
\usepackage[utf8]{inputenc} 
\usepackage[T1]{fontenc} 

\usepackage{url}   
\usepackage{booktabs}  
\usepackage{amsfonts}  
\usepackage{nicefrac}  
\usepackage{microtype}  
\usepackage{xcolor}   
\usepackage{graphicx}
\usepackage{subcaption}
\usepackage{amsmath}
\usepackage{makecell}
\usepackage{nameref}
\usepackage{hyperref}
\usepackage{booktabs}
\usepackage{tabularx}
\usepackage{array}
\usepackage{ragged2e}
\usepackage{fancyhdr}
\usepackage{hyperref}
\usepackage{fancyhdr}
\usepackage{hyperref}

\usepackage{multirow}
\usepackage[utf8]{inputenc} 
\usepackage[T1]{fontenc}    
\usepackage{hyperref}       
\usepackage{url}            
\usepackage{booktabs}       
\usepackage{amsfonts}       
\usepackage{nicefrac}       
\usepackage{microtype}      
\usepackage{xcolor}         
\usepackage{graphicx}
\usepackage{subcaption}
\usepackage{amsmath}

\usepackage{makecell}
\usepackage{multicol}
\usepackage{enumitem}
\usepackage{graphicx}
\newcommand{\heading}[2]{\noindent{\large\itshape S#1. #2}\par\smallskip}
\newcommand{\subentry}[1]{\noindent\hspace{2em}\textbf{#1}\par\smallskip}

\title{\textit{GraphNOSE}: A Graph Transformer in Olfaction }
\begin{document}

\textbf{Abbreviations:} 1-Weisfeiler-Lehman (1-WL) isomorphism test, atoms in molecule localization and delocalization matrices (AIMLDM), area under the receiver operating characteristic curve (AUROC), Chemistry Bidirectional Encoder Representations from Transformers (ChemBERTa), Chemical European Molecular Biology Laborator (ChEMBL), physico-chemical deviance score ($\text{D}_{\text{phys}}$), Geometric Ensemble of Molecules (GEOM), graph convolutional network (GCN), graph isomorphism Network (GIN), graph isomorphism network with edge features (GINE), graph neural network (GNN),  general, powerful and scalable (GPS), graph positional and structural encoder (GPSE), GraphNOSE model utilizing linear transformer ($\text{GraphNOSE}_{\text{lin}}$), GraphNOSE model utilizing standard self-attention  with quadratic complexity ($\text{GraphNOSE}_{\text{trans}}$), curated GoodScents-Leffingwell dataset (GS-LF), graph transformer (GT), kernel density estimation (KDE), K-nearest neighbor (KNN), Laplacian positional encodings (LapPE), limited-memory Broyden-Fletcher-Goldfarb-Shanno (LBFGS), electron localization-delocalization matrix (LDM), $\text{DMPNN}+\text{LDM}$ ($\text{LDM}_{\text{DMPNN}}$), $\text{GCN}+\text{LDM}$ ($\text{LDM}_{\text{GCN}}$) , Molecular ACCess System (MACCS), median absolute deviation for feature $i$ ($\text{MAD}_{i}$), Multi-Layer Perceptron (MLP), Molecular Language Transformer (MolFormer), Message Passing Neural Network (MPNN), out-of-distribution (OOD), open-source counterpart of POM (OpenPOM), Principal Odor Map (POM), Positional and structural encoding (PSE), Receiver Operating Characteristic curve (ROC), Random-walk structural encodings (RWSE), simplified molecular-input line-entry system (SMILES), topological polar surface area (TPSA), XAI (explainable AI) 

\begin{abstract}
 Predicting olfactory qualities from molecular structure is an open problem in chemo-informatics. Although linear models can link molecular features to odor descriptors, they often fail when extrapolating to novel chemical scaffolds, extreme molecular weights, or complex odor mixtures. To address this, we introduce \textit{GraphNOSE}, an open-source graph transformer framework that predicts multi-label odor descriptors from simplified molecular-input line-entry system (SMILES) strings for single molecules and binary mixtures. By integrating positional and structural encodings within a transformer-based graph architecture, \textit{GraphNOSE} achieves strong performance with six times fewer parameters than standard graph neural network (GNN) baseline while consistently outperforming linear models, molecular language model embeddings, molecular fingerprints, and baseline GNNs by an average area under the ROC curve (AUROC) margin of 4.52\% (p < 0.01). \textit{GraphNOSE} achieves an AUROC of 84\% on out-of-distribution compounds (OODs). This exceeds the current state-of-the-art GNN for OOD in olfaction (OpenPOM: 81\%, p < 0.001), and identifies conditions under which linear models empirically fail. Finally, we apply XAI (explainable AI) methods to identify which substructures and molecular features drive odor predictions, yielding insights consistent with chemical intuition and grounded in the model’s learned representations. Together, these results establish \textit{GraphNOSE} as a scalable and interpretable architecture for olfactory prediction that generalizes to structurally distinct compounds underrepresented in current perceptual databases.

\end{abstract}


\section{Introduction}
Predicting human odor perception from molecular structure remains an open and challenging problem in cheminformatics. Olfaction plays a crucial role in human survival and thriving, shaping physiological regulation \cite{tarumi2026physiological,boesveldt2021importance}, feeding behavior \cite{boone2021examining}, and social mating choices \cite{blazing2020odor}, and it underpins the multibillion-dollar food, flavor, and fragrance industries \cite{chen2023food}. Translating physico-chemical properties into perceptual space is particularly complex in olfaction, as structurally similar molecules may evoke qualitatively different odors, revealing a disconnect between molecular structure and perceived odor \cite{sell2006unpredictability} (see Supplementary Material, Figure S1). 

Linear machine learning approaches can nevertheless link molecular features to odor descriptors \cite{schicker2023owsum,keller2017predicting} and show strong predictive power for odorous molecules dominated by isolated functional groups (e.g., aldehydes or thiols) \cite{sanchez2019machine}. Although these models reach a reasonable performance ceiling ( $\approx 80\%$ AUROC) \cite{lee2023principal}, they fail to predict molecules with physico-chemical properties different from those included in the training set. For example, linear classifiers show a marked performance degradation when evaluating structurally novel molecules\cite{lee2023principal, schicker2023owsum}, particularly those with extreme molecular weights. 
In addition, these models fail to generalize to mixtures unseen during training \cite{ravia2020measure}, limiting their utility for ecological olfactory stimuli \cite{arctander2017perfume,kumar2018aromadb,garg2018flavordb,ruddigkeit2014expanding,hastings2016chebi,sharma2022olfactionbase}.

A compounding bottleneck in overcoming these predictive limitations is the fragmentation and lack of standardization in publicly available olfactory databases \cite{parma2026fair}. Odor datasets that combine chemical features and odor descriptors are typically small, often on the order of $10^2$ to $10^3$ molecules \cite{hamel2024pyrfume}, compared to the millions of samples commonly used in other machine learning domains \cite{deng2009imagenet}. Even when aggregation efforts exist, such as Pyrfume library\cite{hamel2024pyrfume}, which integrates olfaction data from multiple sources \cite{arctander2017perfume,kumar2018aromadb,garg2018flavordb,ruddigkeit2014expanding,hastings2016chebi,sharma2022olfactionbase}, annotations remain sparse and inconsistent, particularly with respect to human perceptual descriptors and functional chemical group labels. Despite these limitations, Pyrfume has catalyzed the development of predictive models evaluated on shared benchmarks through initiatives such as the DREAM Olfaction Challenges \cite{keller2017predicting,satarifard2025high}. These efforts enabled the first models to systematically map chemical structure to perceptual space by leveraging predefined, high-dimensional representations of molecular features (i.e., chemical fingerprints \cite{moriwaki2018mordred,morgan1965generation}), thereby facilitating supervised learning of odor descriptors. 

However, while these aggregated repositories significantly increase the number of odorant-label pairs available for model training, they still lack critical dimensions needed to capture the ecological validity of odor perception, most notably stimulus intensity. Odor quality is not a static property of molecular structure but varies with concentration, with some molecules exhibiting qualitative perceptual shifts across intensity ranges. For instance, furanol is described as strawberry-like at low concentrations but as caramel-like at higher concentrations \cite{zabetakis19992}. As a result, descriptors as fixed labels independent of concentration impose a fundamental limitation on current predictive models, which fail to account for the dynamic and context-dependent nature of olfactory perception. \cite{keller2017predicting,lee2023principal,hamel2024pyrfume}.

To overcome the limitations of fingerprint-based odor prediction, researchers initially augmented structural descriptors with mass spectra data \cite{nozaki2016odor}, and subsequently the field shifted from predefined chemical features toward learned representations, in which deep neural networks discover structure-odor mapping directly from the data \cite{zhou2025advances}. Deep learning models have demonstrated strong predictive capability in the field of molecular property prediction by employing graph based approaches, such as message passing neural networks (MPNNs), which represent molecules as atoms (nodes) and bonds (edges) \cite{gilmer2017neural}. Through iterative message passing, atoms exchange information with their neighbors, integrating learned molecular representations to capture structural features and non-linear relationships, enabling more accurate prediction of such properties as drug efficacy, toxicity, and receptor binding \cite{li2022deep}, across domains including quantum chemistry \cite{gilmer2017neural} and drug discovery \cite{wang2025drug}. This transition is substantial: while predefined fingerprints record the presence of isolated chemical substructures, graph-based approaches preserve exact molecular topology. By allowing atoms to iteratively exchange information with their neighbors, the model dynamically learns specific geometric arrangements while explicitly incorporating localized node and edge features, such as formal charge, valence, and structural degree.

In computational olfaction, the most prominent application of these graph-based models is the Principal Odor Map (POM) \cite{lee2023principal} and its open-source counterpart, OpenPOM \cite{aryan_amit_barsainyan_ritesh_kumar_pinaki_saha_michael_schmuker_2023}. These models successfully map 2D molecular graphs of single molecules directly to latent perceptual dimensions. Although recent frameworks like PharmaGNN \cite{liu2026pharmagnn} incorporate static pharmacophore features to capture 3D spatial alignments, all these monomolecular approaches remain critically insufficient for real-world applications, as they do not extend to multi-component odor mixtures. While MPNNs excel at capturing local chemical environments (e.g., functional groups and bonding patterns that impact reactivity), they are constrained by a limited receptive field: information propagates only between neighboring atoms at each step, so many layers are required to propagate information across distant parts of the molecule \cite{rampavsek2022recipe}. 

Deep learning architectures with global receptive fields directly address this aggregation bottleneck by allowing every element to attend to every other in a single step. Such models have proven powerful at capturing long-range dependencies, namely, relationships between elements that are far apart in sequence or space, such as a pronoun and the noun it refers to earlier in a sentence \cite{vaswani2017attention} or two residues that are distant in a protein's sequence yet adjacent once it folds \cite{jumper2021highly}. Global architectures, particularly graph transformers (GTs) \cite{dwivedi2020generalization, ying2021transformers}, have not yet been applied to structure–odor prediction, even though odor descriptors depend on the molecule’s overall geometry and topology rather than on isolated functional groups alone \cite{amoore1964stereochemical}. Even small structural variations (e.g., the exact distance between functional groups or the specific branching of a carbon chain) can dictate receptor binding and evoke different perceptual outcomes \cite{uchida2000odor}. The position of an ester's carbonyl group, for instance, more strongly governs which olfactory receptors it activates than does the identity of the functional group itself \cite{poivet2018functional}. Similarly, while molecules can differ only in the position of a single substituent on the ring, such as carvacrol and thymol, they may trigger the distinctly different odors of oregano and thyme, respectively. This specificity underscores that odor prediction depends on spatial distribution of features across the entire molecule, not on their presence alone. By combining local atomic features into a single global structure, GTs directly mimic how the biological olfactory system integrates receptor signals into the brain. \cite{blazing2020odor}. This spatial awareness is necessary to  predict labels  for both single molecules and mixture odors.

Traditional GTs allow every atom in a molecule to attend to every other atom at once, but they typically treat the molecule as a disconnected “bag of atoms,” ignoring its geometric structure. In other words, they are geometry-blind, failing to capture molecular topology or spatial organization. Positional and structural encodings (PSEs) bridge this gap by providing geometric context: where an atom sits and how its neighborhood is arranged. By injecting these geometric rules directly into the attention mechanism, PSEs force the model to evaluate the molecule as a 3D architecture with explicit topological and geometric information rather than as a mere collection of atoms. Despite several types of PSEs [e.g., random-walk structural encodings (RWSEs) and Laplacian positional encodings (LapPEs) \cite{rampavsek2022recipe}] to select from, no principled guideline optimizes PSE selection for a specific downstream task. Naively concatenating a variety of different PSEs introduces computational overhead (time spend managing rather than doing a task) while often failing to yield proportional performance gains. To bypass the  manual selection of specific PSEs, Cantürk \textit{et al.} \cite{canturk2023graph} proposed the highly transferable Graph Positional and Structural Encoder (GPSE). Rather than relying on domain-specific chemical attributes, GPSE derives highly transferable encodings from graph connectivity alone. These representations can augment any GNN and generalize across graphs of varying sizes, connectivity patterns, and modalities, including both MPNNs and GTs.

Here, we present \textit{GraphNOSE}, an open-source GT-based architecture that combines global attention with GPSE to predict multi-label odor prediction across both single molecules and binary mixtures. We trained and evaluated the model using the largest benchmark dataset available for olfaction (4,983 molecules, 138 descriptors) \cite{lee2023principal, aryan_amit_barsainyan_ritesh_kumar_pinaki_saha_michael_schmuker_2023}. We hypothesized that (a) GPSE will provide more informative positional and structural information than commonly used positional/structural encodings, yielding systematically higher predictive performance on odor-relevant descriptors across attention architectures. Furthermore, we hypothesize that (b) \textit{GraphNOSE}, a GT architecture equipped with GPSE-based spatial encodings, will outperform state-of-the-art MPNNs for molecular odor prediction, as measured by the area under the ROC curve (AUROC) on held-out test data, generalization to out-of-distribution (OOD) molecules, and robustness under reduced training set sizes.  

Our experiments were organized to directly test these two hypotheses and to answer four main questions: 
\begin{itemize}
 \item \textbf{Q1} Which GPSE pretraining corpus provides the most effective inductive bias for \textit{GraphNOSE}, as a model designed for molecular odor prediction?
 
 \item \textbf{Q2} How does \textit{GraphNOSE} compare to existing GNN-based baselines, particularly under extreme physico-chemical out-of-distribution (OOD) conditions and in the low-data regime?
 
 \item \textbf{Q3} Does the learned odor space preserve known hierarchical relationships across odor labels and reflect perceptual similarity beyond mere structural similarity? 
 
 \item \textbf{Q4} How effectively does \textit{GraphNOSE} generalize to binary odor mixtures?

\end{itemize}

Our experiments are organized to directly test our two hypotheses and to answer questions Q1–Q4 through four groups of analyses: (i) GPSE pretraining corpora and inductive bias (Q1); (ii) benchmark comparisons, OOD generalization, and data efficiency (Q2); (iii) structure of the learned odor space and its alignment with perceptual hierarchies (Q3); and (iv) generalization to binary odor mixtures (Q4).

\section{Methods}

\subsection{Dataset and Molecular Graphs}
This work uses a molecular odor dataset derived from the publicly available version of POM \cite{lee2023principal}, namely the OpenPOM release by Barsainyan \textit{et al.} \cite{aryan_amit_barsainyan_ritesh_kumar_pinaki_saha_michael_schmuker_2023}. The dataset contains 4,983 molecules and remains the benchmark for representation learning in olfaction, as each molecule is annotated with up to 138 expert-labeled odor descriptors (e.g., meaty, sweet;  listed in Supplementary Material, Table S1), allowing each odorant to be associated with multiple odor descriptors. Importantly, this is the dataset curated from GoodScents \cite{luebke2019good} and Leffingwell PMP (2001) \cite{leffing}datasets \cite{aryan_amit_barsainyan_ritesh_kumar_pinaki_saha_michael_schmuker_2023} (hereafter jointly referred to as the GS-LF dataset), which were also used to train the original POM model \cite{lee2023principal}, ensuring direct comparability with prior work. In the GS-LF dataset odor descriptor annotations are binary rather than continuous, indicating the presence or absence of a given perceptual quality but not its intensity, and they exhibit a highly imbalanced label distribution (see Supplementary Material, Figure S2a), where highly frequent descriptors like fruity (1,902 molecules) and sweet (1,429 molecules) significantly outnumber rare perceptual qualities such as musk (98 molecules) and garlic (120 molecules), posing significant challenges for model generalization and minority class learning.

\begin{figure}[htbp]
 \centering
\includegraphics[scale=.15]{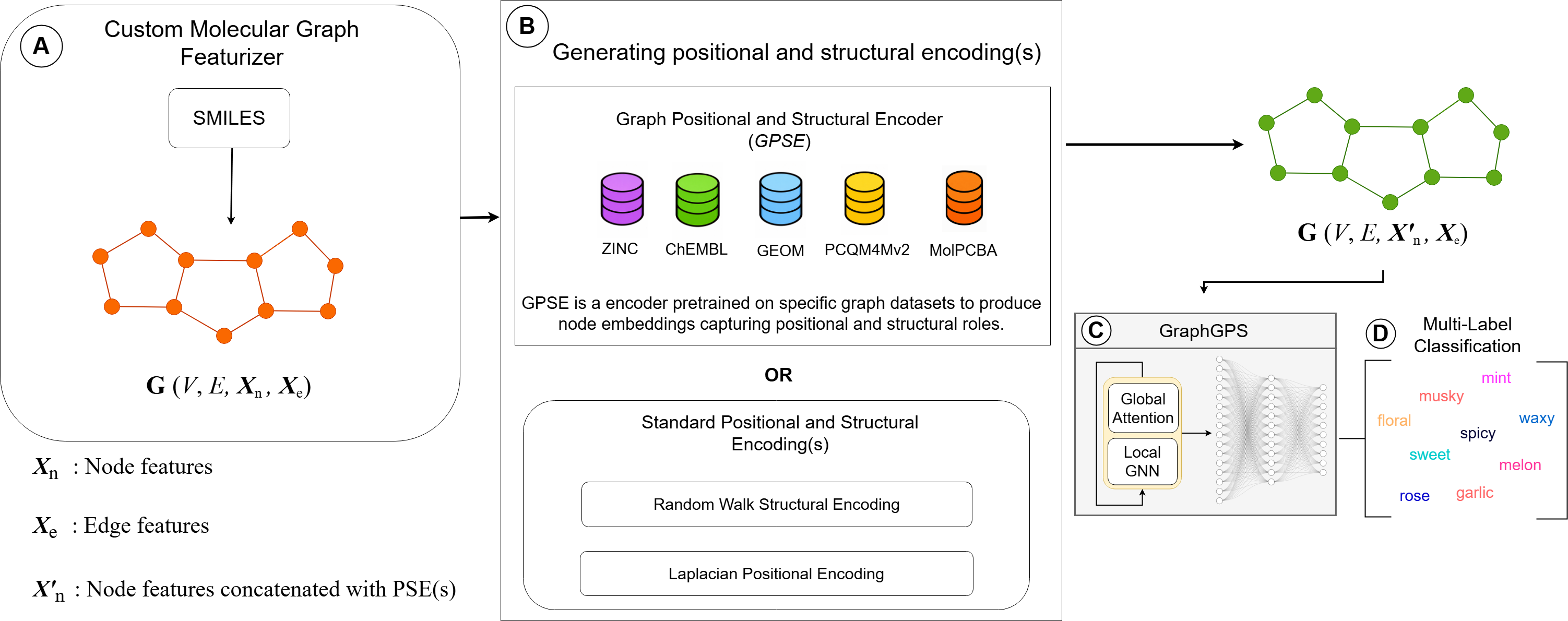}
  \caption{\textbf{Overview of the \textit{GraphNOSE} framework architecture for multi-label molecular odor prediction. (A)} Raw- input SMILES strings are processed through a custom featurizer using RDKit\cite{landrum2016rdkit} and DeepChem \cite{Ramsundar-et-al-2019}, constructing undirected molecular graphs $G(V, E, X_n, X_e)$, where atoms and bonds represent nodes and edges, respectively. \textbf{(B)} The structural graph and node features are augmented using a graph positional and structural encoder (GPSE) pre-trained on diverse molecular and graph datasets (e.g., ZINC, ChEMBL, GEOM, PCQM4Mv2, MolPCBA) or standard baseline (RWSE, LapPE) to generate comprehensive spatial representations ($X'_n$). \textbf{(C)} The enriched graph representations are fed into a hybrid GraphGPS \cite{rampavsek2022recipe} backbone that integrates local message-passing neural networks (MPNNs) for near-neighbor functional group interactions with global self-attention mechanisms for long-range molecular topology. \textbf{(D)} Finally, the model predicts labels for the input molecule from the 138 odor descriptors (e.g., fruity, sweet, musky, garlic).}
 \label{fig:workflow}
\end{figure}
To construct molecular graphs,we employed a custom featurizer on DeepChem \cite{Ramsundar-et-al-2019} (version 2.8.0), utilizing RDKit \cite{landrum2016rdkit} (version 2025.03.6) for SMILES parsing and feature extraction, converting SMILES strings into undirected graphs, where atoms are nodes and bonds are edges. Because we employ the pre-validated OpenPOM dataset\cite{aryan_amit_barsainyan_ritesh_kumar_pinaki_saha_michael_schmuker_2023}, all SMILES strings pass standard RDKit sanitization without any removals. For each node we extract 92 atomic attributes, and for each edge we extract 7 bond attributes (Table \ref{tab:feat}). These features include standard chemoinformatics descriptors such as atomic number, valence, formal charge, hybridization state, aromaticity, and bond order. As the OpenPOM dataset utilizes a strictly non-isomeric SMILES, stereochemical information is not encoded in the model's input features. In other words, molecules that differ only in 3D configuration share identical input representations, and odor descriptors of enantiomers will not be distinguishable. 

The overall \textit{GraphNOSE} pipeline is illustrated in Figure \ref{fig:workflow}: raw SMILES strings are converted into molecular graphs via the custom featurizer; node embeddings are augmented with GPSE-based positional and structural encodings, and the resulting representations are processed by GT to yield multi-label odor predictions.
\begin{table}[htbp]
\centering
\small
\setlength{\tabcolsep}{4pt}
\begin{tabular}{|l|l|}
\hline
\textbf{Feature} & \textbf{Categories} \\
\hline
Atomic number & 1 (H) to 54 (I), UNK \\
\hline
Valence & 0, 1, 2, 3, 4, 5, 6, UNK \\
\hline
Degree & 0, 1, 2, 3, 4, 5, UNK \\
\hline
Formal charge & $-2$, $-1$, 0, 1, 2, UNK \\
\hline
Number of hydrogens & 0, 1, 2, 3, 4, 5, 6, 7, 8, UNK \\
\hline
Hybridization & sp, sp$^{2}$, sp$^{3}$, sp$^{3}$d, sp$^{3}$d$^{2}$, UNK \\
\hline
\textbf{Edge features} & \textbf{Categories} \\
\hline
Bond type & single, double, triple, aromatic, UNK \\
\hline
Is conjugated & true, false \\
\hline
In ring & true, false \\
\hline
\end{tabular}
\caption{Node (atomic) and edge (bond) features used in molecular graphs. All features are one-hot encoded and concatenated into a single bit vector. UNK denotes unknown values and serves as a catch-all category.}
\label{tab:feat}
\end{table}
\subsection{Model}
\label{sec:model}
A fundamental limitation of many GT models is their restricted use of edge information, which can limit their ability to capture local chemical environments \cite{dwivedi2020generalization}. In these architectures, bond features typically only indirectly  influence how information is aggregated across the graph, rather than driving explicit, edge-conditioned updates of node representations. While some GT variants, such as SAN \cite{kreuzer2021rethinking} and Graphormer \cite{ying2021transformers},  inject edge features into the mechanism that determines how strongly nodes weight information from their neighbors, edge attributes still do not directly control the feature updates applied to individual atoms. Rampášek \textit{et al} \cite{rampavsek2022recipe} therefore proposed the General, Powerful and Scalable (GPS) architecture, which combines local message passing with global attention. In GPS, edges play a central role in updating node representations via the message-passing pathway, making the architecture well suited for olfactory modeling, where both local functional groups and global molecular geometry are crucial. 

\textit{GraphNOSE} adopts the GPS architecture \cite{rampavsek2022recipe} to integrate local and global molecular information within each layer. GPS processes node features through two parallel branches: a local MPNN and a global attention module. The MPNN pathway captures localized functional groups and near-neighbor interactions, which are primary drivers for odor prediction\cite{lee2023principal}, while the global self-attention pathway models long-range dependencies across the entire molecular graph.

Formally, at each layer $\ell$, the node features are updated by aggregating the output of the local MPNN and global attention branches. For a graph with $N$ nodes, adjacency matrix $A \in \mathbb{R}^{N \times N}$, edge feature matrix $E$, and the $d_\ell$-dimensional node and edge features $X^\ell \in \mathbb{R}^{N \times d_\ell}$ and $E \in \mathbb{R}^{E \times 7}$, the state transition for the $(\ell+1)$-th layer is defined as

\begin{equation}
 X^{\ell+1} = \text{GPS}^\ell(X^\ell, E, A),
\end{equation}

with the following decomposition:
\begin{align}
 X^{\ell+1}_{M} &= \text{MPNN}^\ell_e(X^\ell, E, A) \\
 X^{\ell+1}_{T} &= \text{GlobalAttn}^\ell(X^\ell) \\
 X^{\ell+1} &= \text{MLP}^\ell(X^{\ell+1}_{M} + X^{\ell+1}_{T})
\end{align}

Here, $\text{MPNN}^\ell_e$ denotes a message-passing network that explicitly uses edge features to model local chemical environments, and $\text{GlobalAttn}^\ell$ denotes a global attention mechanism operating on all node pairs. The fused node embeddings are passed through a two-layer multi-layer perceptron ($\text{MLP}^\ell$). In \textit{GraphNOSE}, edge features are updated and used exclusively within the MPNN component, while global attention operates on node embeddings. This modular design allows the choice of specific MPNN and attention implementations to be tailored to the task.

The core model consists of 4 GPS layers with a hidden dimensionality of 64, utilizing residual connections and layer normalization within each layer to stabilize training. Node features are updated through two parallel branches: a graph isomorphism network with edge features (GINE) for local message passing, where edge attributes are exclusively updated, and single-head attention for global dependencies. Finally, node representations are aggregated via mean pooling and passed through a 512-dimensional MLP classifier for graph-level odor prediction. For detailed model architecture and hyper-parameter configurations, see Supplementary Material, Table S3. We instantiate two architectural variants: GraphNOSE$_{lin}$ uses a linear transformer \cite{choromanski2020rethinking} for global attention, while GraphNOSE$_{trans}$ uses standard self-attention \cite{vaswani2017attention} with quadratic complexity. GraphNOSE$_{lin}$ is primarily employed to avoid the heavy computational overhead associated with standard quadratic attention.

\subsection{Positional and Structural Encoding}
Standard MPNNs are upper-bounded in expressivity by the 1-Weisfeiler-Lehman (1-WL) isomorphism test \cite{xu2018powerful}. To enhance their ability to capture complex molecular topology and to restore structural inductive biases that may be attenuated in global attention, we employ PSEs (positional and structural encodings). These encodings are concatenated with node features, allowing the model to incorporate global structural information into its learned representations. Random-walk structural encodings (RWSE) have proven effective for molecular prediction tasks, while Laplacian positional encodings (LapPE) are well suited to capturing long-range dependencies \cite{rampavsek2022recipe, dwivedi2021graph}. Because no single PSE is universally optimal across tasks, we adopted the graph positional and structural encoder (GPSE) \cite{canturk2023graph} as a pre-trained structural backbone. GPSE is a self-supervised model trained to predict a suite of PSEs, which include RWSE, LapPE \cite{dwivedi2023benchmarking}, heat kernel diagonals, Laplacian eigenvalues, cycle counting, and electrostatic potential encodings. Canturk \textit{et al} \cite{canturk2023graph} showed that latent representations from GPSE can be used to augment both GNNs and GTs as rich positional and structural features.

In \textit{GraphNOSE}, we use the pre-trained GPSE latent embeddings as additional node features and systematically evaluate the impact of different GPSE pre-training corpora on odor prediction. Specifically, we consider GPSE variants pre-trained on high-dimensional biological assays (MolPCBA) \cite{hu2020open}, large-scale quantum properties (PCQM4Mv2) \cite{hu2021ogb}, conformational ensembles (GEOM-Drugs) \cite{axelrod2022geom}, and drug-like chemical space (ZINC/ChEMBL)\cite{gaulton2012chembl,gomez2018automatic}. 
All positional and structural encodings are first projected to a fixed dimensionality via an MLP and then concatenated with node attributes during training. The pre-trained GPSE parameters are kept completely frozen during this process.

\subsection{Training}
\label{sec:training}
We partitioned the GS-LF dataset using an 80:20 train/test split, following the protocol of Lee \textit{et al}\cite{lee2023principal}. Within the 80\% training portion, we reserved 20\% as a validation set for hyperparameter optimization. Due to the multi-label nature of the dataset, we employed iterative stratified splitting to preserve the distribution of label combinations across training, validation, and test sets. No data augmentation techniques (e.g., SMILES enumeration or 3D conformer sampling) are utilized. The model parameters are selected via Bayesian optimization using the Optuna framework \cite{optuna_2019} (the final configuration is summarized in Supplementary Table S3). Specifically, the network is trained using the $Adam$ optimizer with an initial learning rate of $6.82 \times 10^{-4}$ and a batch size of $16$ for a maximum of $150$ epochs. To address label imbalance, we optimize an imbalance-weighted binary cross-entropy loss \cite{tarekegn2021review}, where the contribution of each odor descriptor is weighted inversely to its frequency in the training set. This ensures that rare odor labels are not overwhelmed by more frequent classes. All reported performance metrics, including the area under the ROC curve (AUROC), are calculated as macro-averages to provide an unbiased assessment across the diverse odor label space. To rigorously assess performance differences, paired $t$-tests were performed on seed-matched runs, treating random seeds as paired observations rather than independent samples; $p$-values were reported uncorrected for multiple comparisons.


\subsection{Physico-chemical Out-of-Distribution Splitting}
\label{ood_setting}
Out-of-distribution (OOD) evaluation assesses whether a model can generalize to molecules that differ substantially from those seen during training, rather than merely interpolating within the observed chemical space. To test \textit{GraphNOSE} under pronounced distribution shift, we constructed an OOD split of the GS-LF dataset based on physico-chemical properties. We consider 11 physico-chemical features: molecular weight, logP, hydrogen-bond donors, hydrogen-bond acceptors, topological polar surface area (TPSA), number of rotatable bonds, number of aromatic rings, number of heteroatoms, ring count, number of saturated rings, and number of aliphatic rings. For each molecule, we compute a physico-chemical deviance score  ($\text{D}_{\text{phys}}$) that quantifies its displacement from the dataset median across all properties:

\begin{equation}
\label{eq:extreness}
\text{D}_{\text{phys}} = \frac{1}{11} \sum_{i=1}^{11} \frac{|x_i - \text{median}_i|}{\text{MAD}_i}
\end{equation}
where $\text{MAD}_i$ represents the median absolute deviation for feature $i$. 
Molecules are ranked by $\text{D}_{\text{phys}}$, and we define a 65:35 train/test  split by assigning the most physico-chemically extreme 35\% forming the test set. This is unlike 80:20 iterative stratified split used for training, which randomly samples across the chemical space to preserve odor label balance in training and test sets.

\section{Results and Discussion}

\begin{table}[t]
\centering
\small
\renewcommand{\arraystretch}{1.2}
\caption{\textbf{Different PSE augmentations.} These macro AUROC scores are averaged over five random seeds. Positional and structural encodings (PSEs) are explicitly augmented to the node features to provide topological context to the global attention mechanism. $\text{GraphNOSE}_{\text{lin}}$ uses a linear transformer \cite{choromanski2020rethinking} for global attention, while $\text{GraphNOSE}_{\text{trans}}$ uses standard self-attention \cite{vaswani2017attention} with quadratic complexity. Significance levels ($\ast$ denote paired $t$-test comparisons relative to the $\text{GPSE}_{\text{MolPCBA}}$):
$^{***}p < 0.001$, $^{**}p < 0.01$, and $^{*}p < 0.05$.}
\label{pse-table}
\begin{tabular}{|l|l|l|}
\hline
\textbf{PSE} & $\text{GraphNOSE}_{\text{lin}}$ & $\text{GraphNOSE}_{\text{trans}}$ \\
\hline
RWSE & $87.356 \pm {\scriptstyle 0.15}^{***}$ & $84.474 \pm {\scriptstyle 1.04}^{***}$ \\
\hline
LapPE & $87.796 \pm {\scriptstyle 0.13}^{***}$ & $88.378 \pm {\scriptstyle 0.06}^{***}$ \\
\hline
$\text{GPSE}_{\text{ChEMBL}}$ & $88.864 \pm {\scriptstyle 0.08}$ & $89.084 \pm {\scriptstyle 0.07}$ \\
\hline
$\text{GPSE}_{\text{GEOM}}$ & $88.916 \pm {\scriptstyle 0.05}$ & $89.124 \pm {\scriptstyle 0.13}$ \\
\hline
$\text{GPSE}_{\text{PCQM4Mv2}}$ & $88.930 \pm {\scriptstyle 0.12}$ & $88.990 \pm {\scriptstyle 0.09}$ \\

\hline
$\text{GPSE}_{\text{ZINC}}$ & $88.958 \pm {\scriptstyle 0.18}$ & $89.026 \pm {\scriptstyle 0.10}$ \\
\hline

$\text{GPSE}_{\text{MolPCBA}}$ & $89.024 \pm {\scriptstyle \boldsymbol{0.16}}$ & $89.146 \pm {\scriptstyle 0.08}$ \\
\hline
\end{tabular}
\end{table}
\subsection{GPSE pre-training corpora and inductive bias (Q1)}
\label{optimal_pse}
We investigated the impact of augmenting PSEs such as RWSE \cite{dwivedi2023benchmarking}, LapPE, and learned representations from GPSE \cite{canturk2023graph} to our GT model. We evaluated these encodings with respect to two architectural variant of our model: $\text{GraphNOSE}_{\text{lin}}$ and $\text{GraphNOSE}_{\text{trans}}$. $\text{GraphNOSE}_{\text{lin}}$ utilizes a linear transformer (Performer \cite{choromanski2020rethinking}) as the global attention, whereas, $\text{GraphNOSE}_{\text{trans}}$ uses self-attention \cite{vaswani2017attention} for the global attention mechanism, which has quadratic complexity.

Table \ref{pse-table} demonstrates that the GPSE trained on varied chemical datasets consistently outperforms explicitly constructed PSEs. While \textit{GraphNOSE} achieves its highest numerical performance with MolPCBA pre-training, the framework exhibits robust performance across all evaluated pre-training corpora, indicating that \textit{GraphNOSE} is not overly sensitive to the selection of the underlying training domain.
Although we hypothesize that this minor numerical advantage may stem from a favorable alignment between MolPCBA biological assay targets and olfactory receptor-binding mechanisms, the differences across corpora remain largely statistically indistinguishable.

From a theoretical perspective, linear transformers like Performer \cite{choromanski2020rethinking} are characterized by lower expressivity compared to self-attention mechanisms. Across most positional and structural encodings, $\text{GraphNOSE}_{\text{lin}}$ and $\text{GraphNOSE}_{\text{trans}}$ perform equivalently, showing negligible performance gaps ($\le 0.12\%$) with overlapping standard deviations. An exception is observed with RWSE, where $\text{GraphNOSE}_{\text{trans}}$ degrades markedly in performance relative to $\text{GraphNOSE}_{\text{lin}}$
(Table \ref{pse-table}, line 1). The performance degradation may be attributed to the attention collapse in RWSE, where the model pays more attention to the local structural features, and this leads to an attention matrix, which is often sparse. This does not persist in \textit{Performers}\cite{choromanski2020rethinking}, which handles the attention matrix by the low-rank kernel approximation, or in LapPE, which constructs it from a graph Laplacian matrix that encodes global graph geometry.

\begin{table}[ht]
\centering
\caption{\textbf{Baselines}. Models are grouped into descriptor-based methods, LLM embedding approaches, standard GNN architectures, other GNN variants, and the proposed method. Linear models using the limited-memory Broyden-Fletcher-Goldfarb-Shanno (LBFGS) solver are deterministic and therefore do not report standard deviations, paired $p$-values, or effect sizes. Significance levels and effect sizes ($\ast$ denote paired $t$-test comparisons and Cohen's $d$ relative to the $\text{GraphNOSE}_{\text{lin}}$ baseline): $^{***}p < 0.001$, $^{**}p < 0.01$, and $^{*}p < 0.05$. LR, logistic regression; RF, random forest.}
\label{tab:model_comparison}
\resizebox{\textwidth}{!}{%
\begin{tabular}{|c|l|c|c|c|}
\hline
\textbf{Class} & \textbf{Model} & \textbf{ROC (\%)} & \textbf{$p$-value} & \textbf{Cohen's $d$} \\
\hline
\multirow{5}{*}{Linear [logistic regression (LR)]}
& LR (TopTorsion) \cite{nilakantan1987topological} & $80.82$ & -- & -- \\
\cline{2-5}
& LR (RDKit) \cite{landrum2016rdkit}& $82.05$ & -- & -- \\
\cline{2-5}
& LR (AtomPair) \cite{carhart1985atom} & $82.83 $ & -- & -- \\
\cline{2-5}
& LR (Morgan\_ECFP4) \cite{rogers2010extended} & $83.56 $ & -- & -- \\
\cline{2-5}
& LR (MACCS) \cite{durant2002reoptimization}& $84.76 $ & -- & -- \\
\hline
\multirow{2}{*}{LLM embedding}
& ChemBERTa \cite{chithrananda2020chemberta}& $84.78 \pm {\scriptstyle 0.04}^{***}$ & $4.93\times10^{-7}$ & $26.40$ \\
\cline{2-5}
& MoLFormer \cite{wu2023molformer}& $84.97 \pm {\scriptstyle 0.09}^{***}$ & $3.12\times10^{-8}$ & $52.66$ \\
\hline
\multirow{3}{*}{Fingerprint}
& RF-bFP \cite{rogers2010extended}& $85.00 \pm {\scriptstyle 0.08}^{***}$ & $9.67\times10^{-7}$ & $22.31$ \\
\cline{2-5}
& RF-cFP \cite{rogers2010extended} & $85.01 \pm {\scriptstyle 0.09}^{***}$ & $1.81\times10^{-6}$ & $19.08$ \\
\cline{2-5}
& RF-Mordred \cite{moriwaki2018mordred} & $85.81 \pm {\scriptstyle 0.15}^{***}$ & $2.27\times10^{-6}$ & $18.01$ \\
\hline
\multirow{3}{*}{GNN}
& $\text{LDM}_{\text{DMPNN}}$ \cite{D5DD00224A}& $86.60 \pm {\scriptstyle 0.25}^{***}$ & $8.34\times10^{-5}$ & $7.28$ \\
\cline{2-5}
& $\text{LDM}_{\text{GCN}}$ \cite{D5DD00224A}& $86.91 \pm {\scriptstyle 0.15}^{***}$ & $3.73\times10^{-5}$ & $8.92$ \\
\cline{2-5}
& OpenPOM \cite{OpenPOM}& $87.14 \pm {\scriptstyle 0.57}^{**}$ & $2.77\times10^{-3}$ & $2.94$ \\
\hline
\multirow{2}{*}{Graph transformer}
& $\text{GraphNOSE}_{\text{lin}}$ & $89.02 \pm {\scriptstyle 0.16}$ & -- & -- \\
\cline{2-5}
& $\text{GraphNOSE}_{\text{trans}}$ & $89.15 \pm {\scriptstyle 0.08}$ & -- & -- \\
\hline
\end{tabular}
}
\end{table}

\subsection{Benchmark comparisons, OOD generalization, and data efficiency (Q2)}
\label{baselines}
To compare the predictive performance of \textit{GraphNOSE}, we benchmarked it against various state-of-the-art GNNs, chemical language models, and classical chemoinformatics fingerprint-based methods.

For traditional machine learning baselines, we evaluated both linear and non-linear architectures. Linear baselines utilized logistic regression paired with traditional 2D molecular fingerprints [Morgan (Extended-Connectivity Fingerprint, diameter 4),  \cite{morgan1965generation,rogers2010extended} RDKit, MACCS \cite{durant2002reoptimization}, AtomPair \cite{carhart1985atom}, and TopTorsion \cite{nilakantan1987topological}]  to establish the foundational performance boundary. For our advanced non-linear baselines, we utilized random forests, LLM embeddings augmented by multi-layer perceptrons (MLPs), and inherently non-linear graph neural networks (GNNs). To ensure that the performance comparison was fair, all baseline models were trained and evaluated using the exact same GS-LF dataset splits as for training and the extreme physico-chemical OOD split as described above (see \nameref{ood_setting} in Methods, above). All data splits and baseline evaluation scripts are publicly accessible from our \href{https://github.com/CSIO-FPIL/GraphNOSE}{GitHub repository}.


\paragraph{LLM embedding-based methods} As representations derived from pre-trained LLMs align closely with perceptual representation of odors \cite{taleb2024can}, we incorporated embeddings of chemistry-informed LLM models ChemBERTa (768 dimensions) \cite{chithrananda2020chemberta} and MoLFormer (768 dimensions) \cite{ross2022large} in the baselines. We fed the embeddings into a robust MLP classifier optimized via Optuna \cite{optuna_2019} and trained with the same weighted loss function to have fair comparision with our model.

\paragraph{GNN-based methods} We compared \textit{GraphNOSE} with several inherently non-linear, state-of-the-art architectures optimized specifically for olfaction. Primarily we compared it with OpenPOM, \cite{aryan_amit_barsainyan_ritesh_kumar_pinaki_saha_michael_schmuker_2023} the publicly available version of the model proposed by Lee \textit{et al}\cite{lee2023principal}. OpenPOM uses a custom featurizer and utilizes MPNN to map chemical structures to perceptually meaningful latent space. We compared the model with two variants of a GNN that utilizes the atoms in molecule localization and delocalization matrices (AIMLDM) \cite{11sumar2015aimldm} approach for odor prediction and incorporates quantum-informed electron density distributions into the graph’s adjacency matrix: \cite{D5DD00224A} $\text{GCN}+\text{LDM}$ ($\text{LDM}_{\text{GCN}}$) and $\text{DMPNN}+\text{LDM}$ ($\text{LDM}_{\text{DMPNN}}$). 

\begin{table}[ht]
\centering
\small
\caption{\textbf{Parameter counts for single molecule odor prediction models}, reported in thousands (K) and millions (M).}
\label{tab:parameter_comparisons}
\begin{tabular}{|l|c|}
\hline
\textbf{Model} & \textbf{Parameters} \\
\hline
OpenPOM & 2M \\
\hline
$\text{LDM}_{\text{DMPNN}}$ & 674K \\
\hline
$\text{LDM}_{\text{GCN}}$ & 671K \\
\hline
\textit{GraphNOSE} & 314K \\
\hline

\end{tabular}
\end{table}
The results for model performances on the held-out test dataset are displayed in Table \ref{tab:model_comparison}. To establish a foundational performance boundary, linear models achieved a maximum AUROC of 84.76\% (logistic regression with MACCS). \textit{GraphNOSE} significantly outperformed OpenPOM (paired $t$-test, $p = 0.0028$, $d = 2.94$), $\text{LDM}_{\text{GCN}}$ ($p = 3.73 \times 10^{-5}$, $d = 8.92$), and $\text{LDM}_{\text{DMPNN}}$ ($p = 8.34 \times 10^{-5}$, $d = 7.28$), while using only 314K parameters, compared to OpenPOM (2M parameters) and LDM baselines ($\approx 670$K parameters)
(Table \ref{tab:parameter_comparisons}). Though the performance of GNNs may appear marginal in absolute terms compared to fingerprint-based methods, these gains are significant when considering the high-dimensional sparsity and long-tailed label distribution inherent in the GS-LF dataset (see Supplementary Figure S2a). The $\text{GraphNOSE}_{\text{lin}}$ and $\text{GraphNOSE}_{\text{trans}}$ results in Table \ref{tab:model_comparison} are derived using $\text{GPSE}_{\text{MolPCBA}}$ augmentation, as evaluation across pre-trained corpora identified it as yielding the highest numerical performance (Table \ref{pse-table}). The same was used in subsequent experiments.
\begin{figure}[htpb]
\centering

\includegraphics[width=0.6\textwidth]{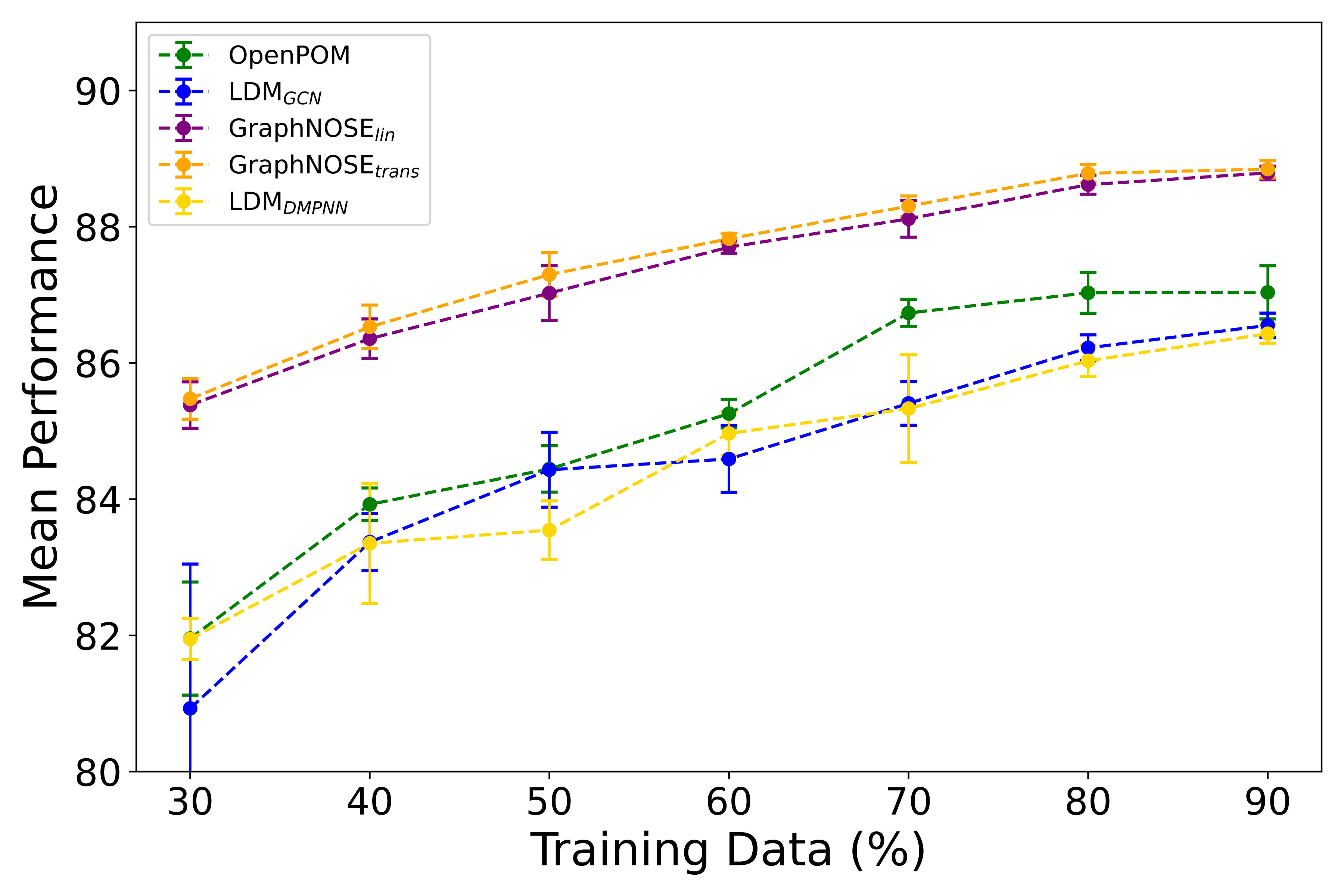}
\caption{\textbf{Impact of training set size on model performance:} Mean predictive performance on the held-out test set as a function of the training data proportion. Error bars represent the standard deviation across 5 random seeds. While all architectures exhibit performance gains as the training set size increases, both \textit{GraphNOSE} variants consistently outperform baseline GNNs (OpenPOM and LDM variants) across all sampling rates, demonstrating superior data efficiency even in low-data regimes.}
\label{fig:model_comparison}
\end{figure}
\paragraph{\textit{GraphNOSE} outperforms GNNs in low-data regimes} We evaluated the performance of our model on data-constrained conditions. This can be visualized by measuring performance on the held-out test dataset (Figure \ref{fig:model_comparison}). Both \textit{GraphNOSE} variants ($\text{GraphNOSE}_{\text{lin}}$ and $\text{GraphNOSE}_{\text{trans}}$) consistently outperformed other GNN baselines across the entire spectrum of data availability (30--90\%, $p < 0.05$ to $p < 0.001$; see Supplementary Tables S4--S6).
The performance margin is largest in the 30-50\%
 regime, where the baseline performance diminishes. We hypothesize that the parameter efficiency of the proposed model, coupled with the strong structural priors provided by GPSE, facilitates this generalization. 
\begin{table}[htb]
\centering
\small
\caption{\textbf{Out-of-distribution (OOD) training results for molecule odor prediction across five random seeds}. As linear models using the LBFGS solver are deterministic, they do not report standard deviations, paired $p$-values, or effect sizes. The best-performing score is highlighted in bold. LR, logistic regression. Significance levels ($\ast$ denote paired $t$-test comparisons relative to the $\text{GraphNOSE}_{\text{lin}}$ baseline):
$^{***}p < 0.001$, $^{**}p < 0.01$, and $^{*}p < 0.05$.}
\label{tab:ood_training}
\begin{tabular}{|l|c|c|c|}
\hline
\textbf{Model} & \textbf{AUROC (\%)} & \textbf{$p$-value} & \textbf{Cohen's $d$} \\
\hline
$\text{GraphNOSE}_{\text{trans}}$ & $\boldsymbol{84.34 \:\pm\:} {\scriptstyle \boldsymbol{ 0.18}}$ & -- & --\\
\hline
$\text{GraphNOSE}_{\text{lin}}$ & $84.26 \pm {\scriptstyle 0.33}$ & -- & -- \\
\hline
OpenPOM & $80.84 \pm {\scriptstyle 0.35}^{***}$ & $4.39\times10^{-7}$ & $27.19$ \\
\hline
$\text{LDM}_{\text{GCN}}$ & $79.13 \pm {\scriptstyle 0.44}^{***}$ & $1.52\times10^{-5}$ & $11.18$ \\
\hline
$\text{LDM}_{\text{DMPNN}}$ & $79.03 \pm {\scriptstyle 0.73}^{***}$ & $5.63\times10^{-5}$ & $8.04$ \\
\hline
LR (MACCS) & 78.46 & -- & -- \\
\hline
LR (Morgan\_ECFP4) & 77.31 & -- & -- \\
\hline
LR (TopTorsion) & 75.77 & -- & -- \\
\hline
LR (AtomPair) & 75.47 & -- & -- \\
\hline
LR (RDKit) & 74.52 & -- & -- \\
\hline
\end{tabular}
\end{table}


\paragraph{\textit{GraphNOSE} outperforms GNNs in extreme out-of-distribution (OOD) setting} \quad 

To evaluate the generalization capacity of \textit{GraphNOSE}, performance was tested under a severe distribution shift using the physico-chemical-based split of the GS-LF dataset (see \nameref{ood_setting}, above). Rather than evaluating with molecules structurally similar to the training data, the models were tasked with predicting odor labels for the most physico-chemically extreme compounds. For reproducibility, we provide the exact OOD training and test splits in the \href{https://github.com/CSIO-FPIL/GraphNOSE/tree/main/data}{GitHub repository}.
Despite this challenging setup, both \textit{GraphNOSE} variants outperform the baseline GNNs by an average margin of $4.65\%$ ($p < 0.001$, paired t-test). For context, the strongest linear baseline performance on this extreme split topped out at an AUROC of 78.46\% (logistic regression with MACCS; see Table \ref{tab:ood_training}). \textit{GraphNOSE}'s superior performance can be hypothesized to stem from the global attention mechanism that captures the long-range dependencies between distant functional groups that remain consistent even when the local scaffold changes (see Table \ref{tab:ood_training}). Alternative explanations for this performance might include the robust structural priors injected via $\text{GPSE}$ and the parameter-efficient modeling ($\sim$314K parameters).

\begin{figure}[htpb]
 \centering
  \includegraphics[scale=0.21]{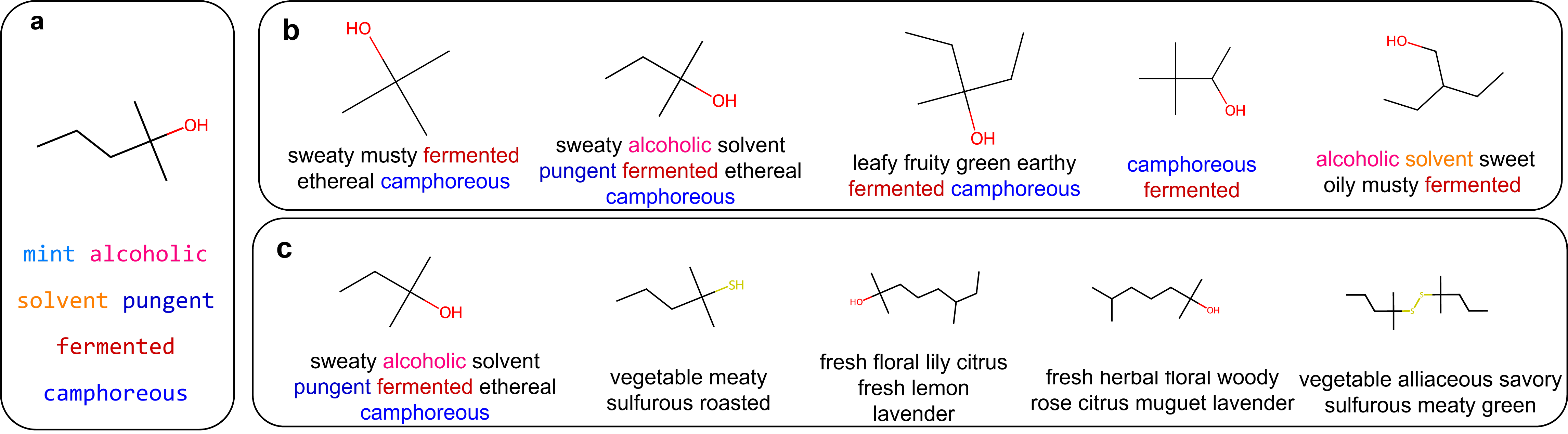}
 \caption{\textbf{Nearest neighbor retrieval for a reference molecule.}\textbf{a)} Reference molecule \textbf{b)} Top five nearest neighbors retrieved using cosine similarity on $\textit{GraphNOSE}$ continuous embeddings. \textbf{c)} Top five nearest neighbors retrieved using Tanimoto distance on bit-based Morgan fingerprints (serving as the established baseline in chemo-informatics\cite{jasial2016activity}).}
 \label{fig:nearest-neighbour}
\end{figure}

\subsection{Structure of the learned odor space and its alignment with perceptual hierarchies (Q3)}
\label{embed_space}
We examined the learned space of $\text{GraphNOSE}_{\text{lin}}$ (chosen for its computational efficiency via linear attention while yielding performance comparable to $\text{GraphNOSE}_{\text{trans}}$ as detailed in Table \ref{pse-table}) by employing the activations from the penultimate layer of the model (512 dimensions). This analysis follows the experiments by Lee \textit{et al}\cite{lee2023principal}.

\paragraph{Perceptually similar molecules cluster together} To evaluate the representational quality of the learned embedding space, we assessed whether local proximity also corresponds to perceptual similarity. Five nearest neighbors were identified from the learned space of $\text{GraphNOSE}_{\text{lin}}$ and compared with those neighbors identified via Tanimoto similarity on bit-based Morgan fingerprints (Figure \ref{fig:nearest-neighbour}). To provide a descriptive diagnostic of the local neighborhood structure, a K-nearest neighbor (KNN) classifier ($k=50$) was trained on a single evaluation split of the penultimate-layer embeddings of $\text{GraphNOSE}_{\text{lin}}$ and compared against bit-based Morgan fingerprints \cite{morgan1965generation}. This descriptive evaluation yields an AUROC of 86.91\% for $\text{GraphNOSE}_{\text{lin}}$ versus 79.73\% for Morgan fingerprints, illustrating how the embedding space efficiently captures perceptual similarity in local neighborhoods. 

\begin{figure}[tbp]
 \centering
 \begin{subfigure}{0.48\textwidth}
  \centering
  \includegraphics[width=\textwidth]{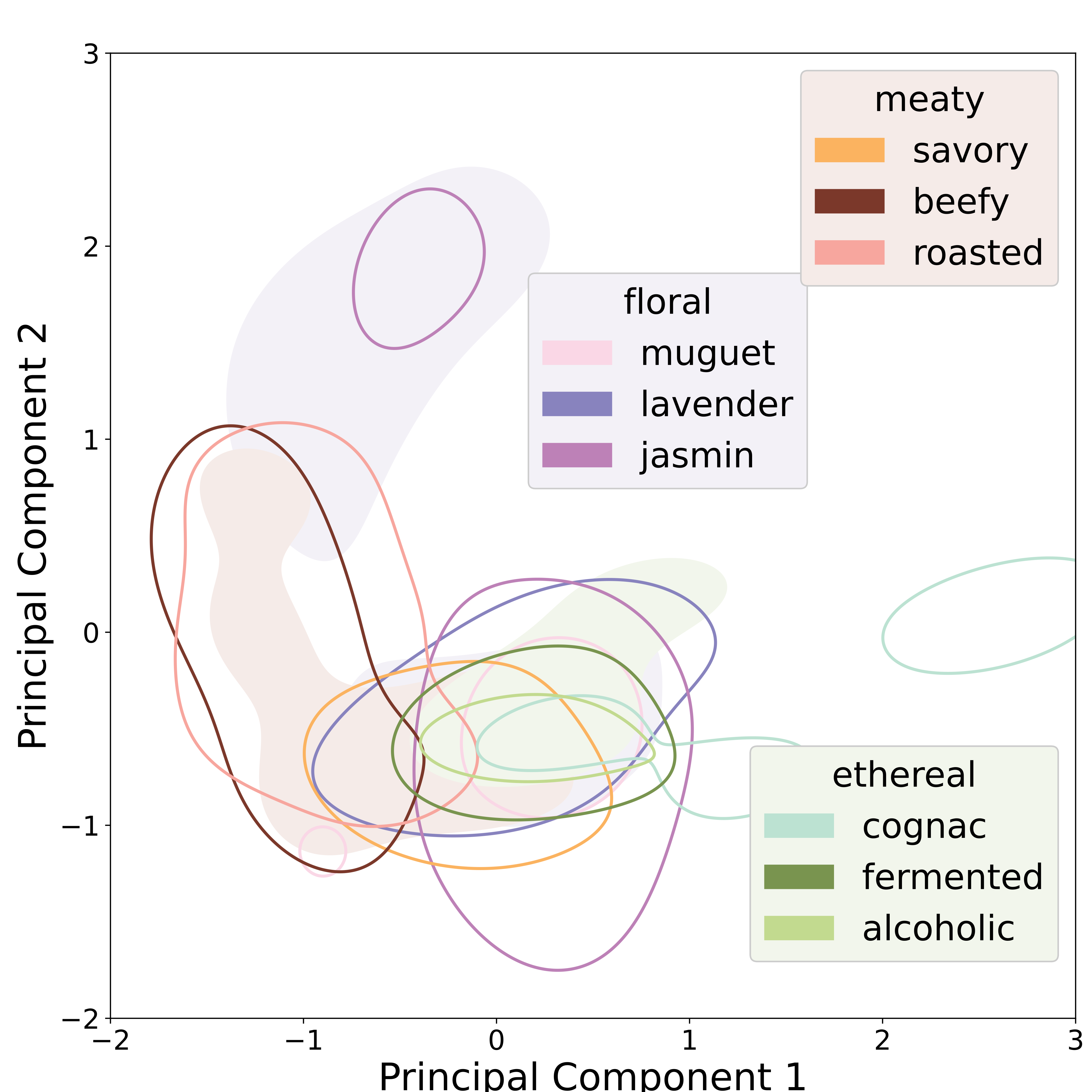}
  \caption{}
  \label{fig:fp_kde}
 \end{subfigure}%
 \hfill%
 \begin{subfigure}{0.48\textwidth}
  \centering
  \includegraphics[width=\textwidth]{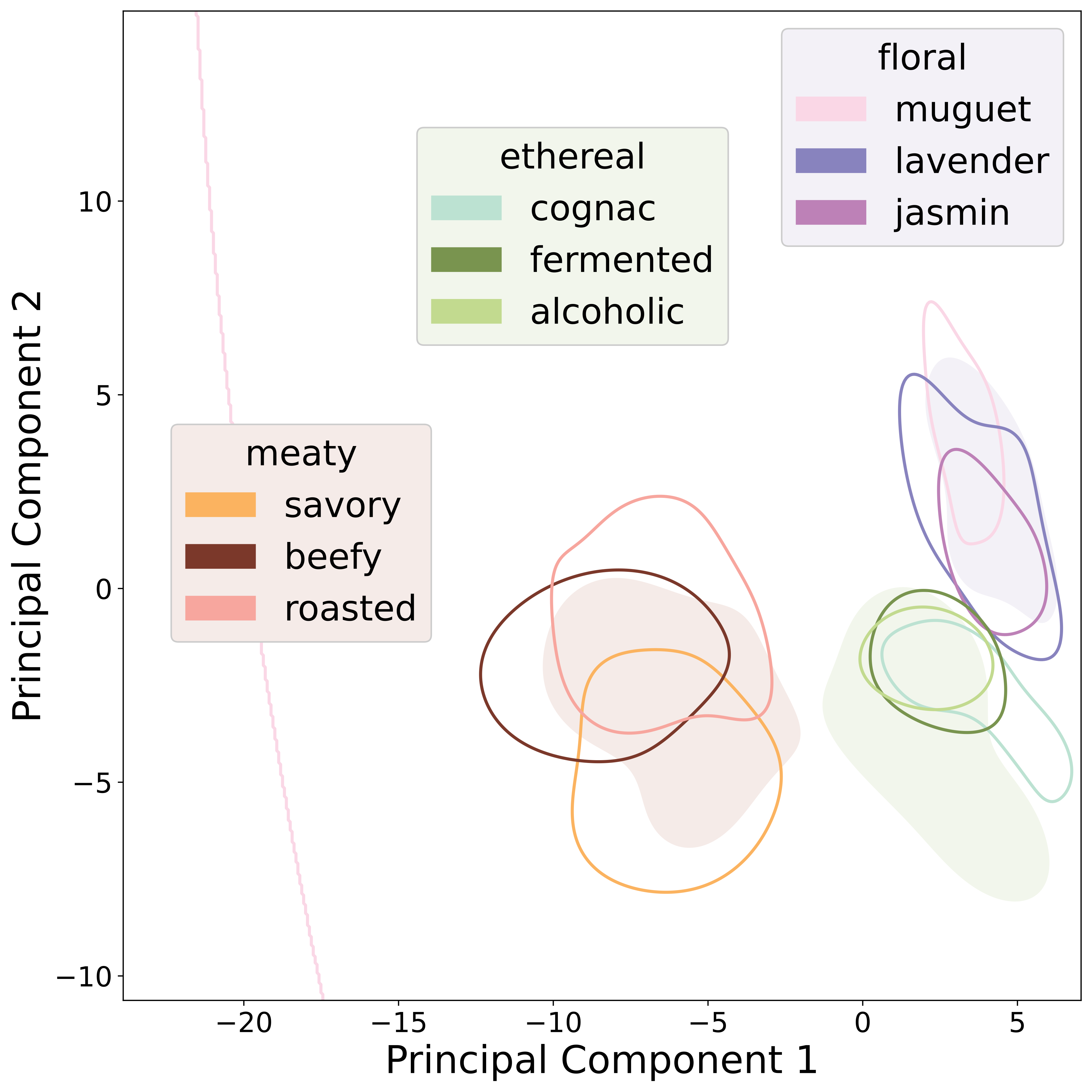}
  \caption{}
  \label{fig:labels_kde}
 \end{subfigure}
  \hfill%
 \begin{subfigure}{0.48\textwidth}
  \centering
  \includegraphics[width=\textwidth]{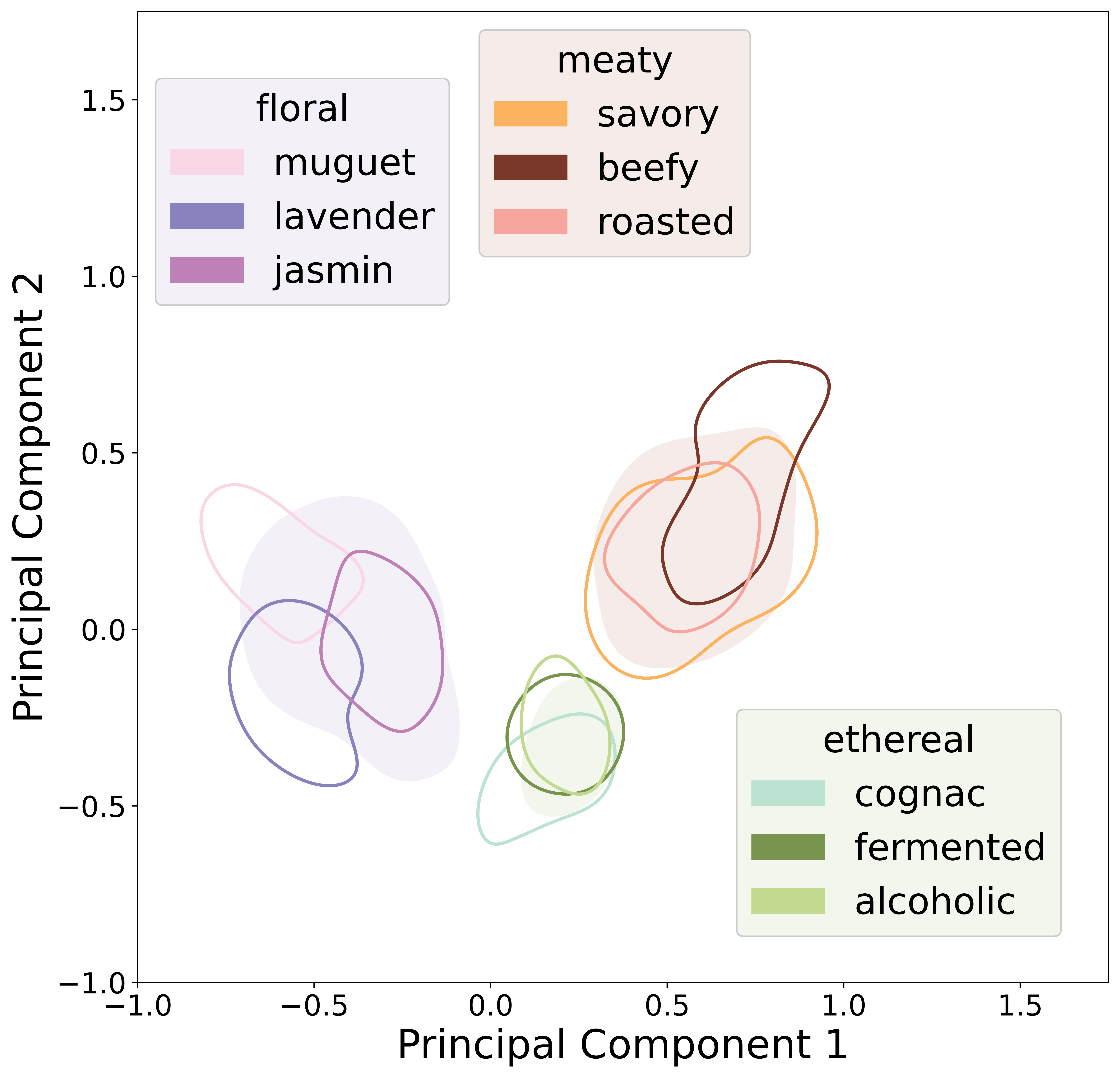}
  \caption{}
  \label{fig:POM_kde}
 \end{subfigure}
 \caption{\textbf{Analysis of odor space:} Kernel density estimation (KDE) plots for Morgan fingerprint \textbf{(a)} and predicted labels by OpenPOM \textbf{(b)} and GraphNOSE \textbf{(c)} Shaded and contoured areas are kernel density estimates of the distribution of labeled data. Colors denote odor categories, with shaded areas for broad classes and contours for specific descriptors (e.g., lavender, roasted). }
 \label{fig:combined}
\end{figure}
\paragraph{Odor-label semantics are preserved in embedding space} 
The presence of neighborhoods consisting of molecules exhibiting similar odor profiles is necessary but not sufficient for an optimal odor space. The distribution of molecules in the embedding space must indicate the logical hierarchies between the odor labels. The learned space should preserve logical hierarchies and semantic relationships between odor labels. Figure \ref{fig:fp_kde} illustrates the kernel density estimation (KDE) plots derived from Morgan fingerprints, where the boundaries between different odor categories are largely indistinct. This highlights the limitations of purely structural representations in capturing perceptual organization. In contrast, the KDE of the true labels (see Supplementary Material, Figure S3), serves as a benchmark, demonstrating both the separation of dissimilar odors (e.g., floral vs. meaty) and the hierarchical relationships within categories (e.g., jasmine, lavender, and muguet within floral). Figure \ref{fig:POM_kde} presents the embeddings learned by our model, where the global odor space exhibits clear separation between distinct odor families while also preserving their hierarchical structure. In particular, \textit{GraphNOSE} successfully distinguishes between broad perceptual classes while simultaneously capturing fine-grained relationships among subcategories. These results suggest that \textit{GraphNOSE} learns a perceptually relevant latent representation that more closely reflects human odor perception than conventional fingerprint-based embeddings.

\begin{figure}[ht]
\centering
\begin{subfigure}[b]{0.30\textwidth}
 \centering
 \includegraphics[width=\textwidth]{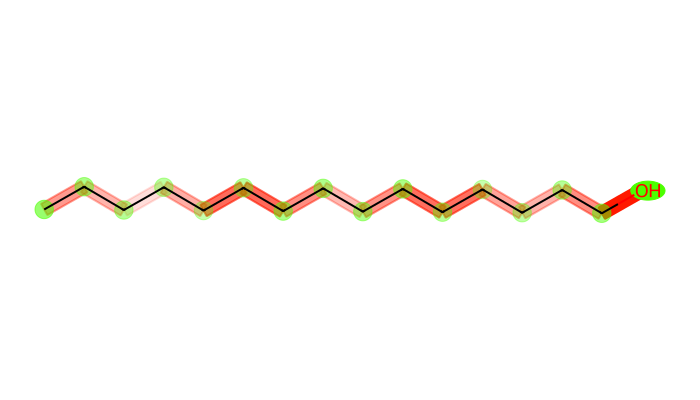}

 \caption{Alcohol}
 \label{fig:explain_alcohol}
\end{subfigure}
\begin{subfigure}[b]{0.30\textwidth}
 \centering
 \includegraphics[width=\textwidth]{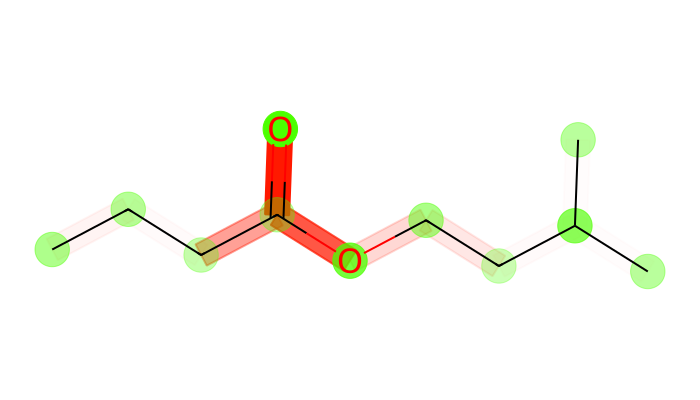}
 \caption{Fruity}
 \label{fig:explain_fruity}
\end{subfigure}
\begin{subfigure}[b]{0.30\textwidth}
 \centering
 \includegraphics[width=\textwidth]{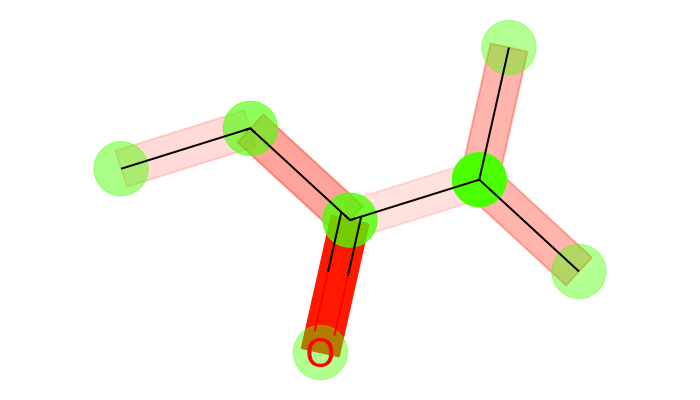}
 \caption{Ketonic}
 \label{fig:explain_ketonic}
\end{subfigure}


\begin{subfigure}[b]{0.30\textwidth}
 \centering
 \includegraphics[width=\textwidth]{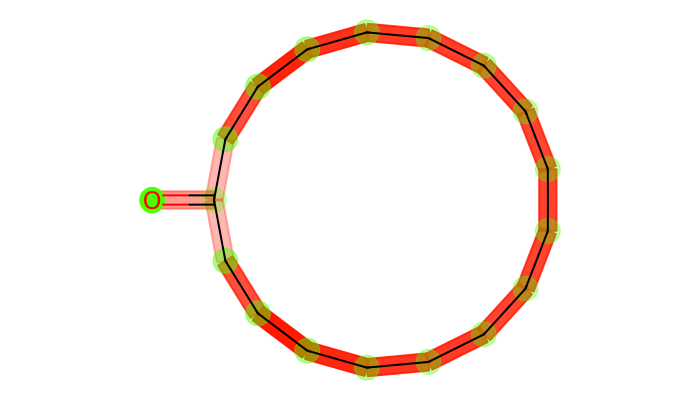}
 \caption{Musk}
 \label{fig:explain_musk}
\end{subfigure}
\begin{subfigure}[b]{0.30\textwidth}
 \centering
 \includegraphics[width=\textwidth]{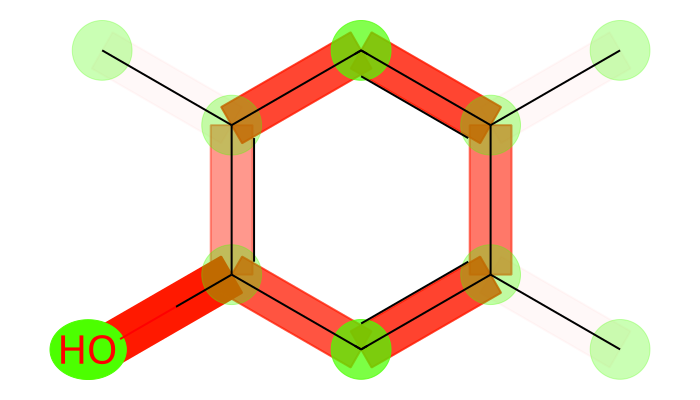}
 \caption{Phenolic}
 \label{fig:explain_phenolic}
\end{subfigure}
\begin{subfigure}[b]{0.30\textwidth}
 \centering
 \includegraphics[width=\textwidth]{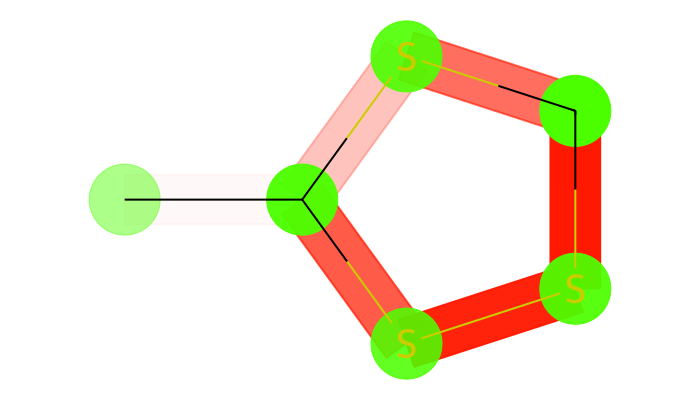}
 \caption{Sulfurous}
 \label{fig:explain_sulfurous}
\end{subfigure}

\caption{\textbf{Explainability through integrated gradients \cite{sundararajan2017axiomatic} of different odor descriptors.}The visualizations highlight the molecular features most important for predicting each odor category. Green highlights atom-level contributions; red highlights bond-level contributions. Node and edge importance are visualized by the opacity and are proportional to the values. For more examples in each category, see Supplementary Material, Figure S4.}

\label{fig:explainability}
\end{figure}
\subsection{Explainability: Structure-Odor Relationship}
\label{sec:explain}
Integrated gradients \cite{sundararajan2017axiomatic} analysis reveals that the GT model consistently focuses on canonical odor-bearing functional groups for each class. Figure \ref{fig:explain_alcohol} shows that alcoholic odors show high attribution on hydroxyl (O–H and C–O) groups, whereas Figure \ref{fig:explain_phenolic} shows that phenolic odors emphasize the phenolic OH and adjacent aromatic carbons. The molecules for which the model prediction was sulfurous odors highlight sulfur atoms and adjacent carbon chains (Figure \ref{fig:explain_sulfurous}). Ketonic odors emphasize carbonyl (C=O) groups and $\alpha$-carbons in conjugated systems (Figure \ref{fig:explain_ketonic}). The musk odor label highlights the importance of constituent called muscone, a macrocyclic ketone (Figure \ref{fig:explain_musk}). Fruity odor labels revealed that the ester groups were key contributors (Figure \ref{fig:explain_fruity}). These patterns are consistent with chemical intuition regarding the functional groups that influence olfaction. This is consistent with the model learning chemically meaningful structure–odor relationships. The insights gained increase confidence in model predictions. 
\begin{table}[htbp]
\centering
\small
\renewcommand{\arraystretch}{1.2}
\caption{\textbf{Results for binary odor mixture prediction.} Scores are macro AUROC scores averaged over five random seeds. Linear models (LR, logistic regression) are deterministic and therefore do not report standard deviations, paired $p$-values, or effect sizes. Significance levels ($\ast$ denote paired $t$-test  comparisons relative to $\text{GPSE}_{\text{MolPCBA}}$):
$^{***}p < 0.001$, $^{**}p < 0.01$, and $^{*}p < 0.05$.}
\label{tab:mix}
\begin{tabular}{|l|l|c|c|c|}
\hline
\textbf{Model} & \textbf{PSE / Features} & \textbf{ROC (\%)} & \textbf{$p$-value} & \textbf{Cohen's $d$} \\
\hline
 \multirow{7}{*}{$\text{GraphNOSE}_{\text{lin}}$} & $\text{GPSE}_{\text{MolPCBA}}$ & $77.96 \pm {\scriptstyle 0.58}$ & -- & -- \\
 \cline{2-5}
 & $\text{GPSE}_{\text{ChEMBL}}$ & $77.84 \pm {\scriptstyle 0.54}$ & $0.778$ & $0.14$ \\
 \cline{2-5}
 & $\text{GPSE}_{\text{pcqmv2}}$ & $77.64 \pm {\scriptstyle 0.99}$ & $0.349$ & $0.47$ \\
 \cline{2-5}
 & $\text{GPSE}_{\text{zinc}}$ & $77.58 \pm {\scriptstyle 0.67}$ & $0.396$ & $0.42$ \\
 \cline{2-5}
 & LapPE & $77.54 \pm {\scriptstyle 1.24}$ & $0.601$ & $0.25$ \\
 \cline{2-5}
 & $\text{GPSE}_{\text{Geom}}$ & $77.24 \pm {\scriptstyle 0.63}^{*}$ & $0.038$ & $1.36$ \\
 \cline{2-5}
 & RWSE & $74.66 \pm {\scriptstyle 1.13}^{**}$ & $5.00\times10^{-3}$ & $2.50$ \\
\hline
MPNN-GNN & - & $76.64 \pm {\scriptstyle 0.06}^{**}$ & $8.31\times10^{-3}$ & $2.17$ \\
\hline
LR (Morgan\_ECFP4) & - & 74.05 & -- & -- \\
\hline
LR (TopTorsion) & - & 73.74 & -- & -- \\
\hline
LR (MACCS) & - & 72.91 & -- & -- \\
\hline
LR (AtomPair) & - & 72.87 & -- & -- \\
\hline
LR (RDKit) & - & 71.28 & -- & -- \\
\hline
\end{tabular}
\end{table}

\subsection{Generalization to binary odor mixtures (Q4)}
\label{mix_pred}
Odor mixing is intricate rather than being a simple additive process. It is characterized by complex interactions where existing odors are often suppressed and entirely new olfactory qualities emerge \cite{sisson2025deep}. We evaluated the capability of $\text{GraphNOSE}_{\text{lin}}$ to generalize to binary mixtures. To process these binary mixtures computationally, the SMILES strings of the two constituent molecules were concatenated using a standard dot (.) separator. This generates a single, disconnected molecular graph as input. We adopted exactly the same "graph carving" train/test split (50:50) proposed by Sisson \textit{et al.} \cite{sisson2025deep}, which partitions the dataset on the basis of unseen molecular structures and not unseen combinations of known molecules. This strict separation (44,000 training pairs and 40,000 test pairs) prevents data leakage and learns representations for novel chemical structures with 74 odor labels. The MPNN framework proposed by Sisson \textit{et al.} \cite{sisson2025deep} outperformed other baselines (including GIN, support vector machine (SVM), and random forest) on this specific split. We benchmarked our model directly against this state-of-the-art MPNN baseline and linear models (Table \ref{tab:mix}). We used the same optimal hyperparameters chosen for the single-molecule model, increasing only the training epochs that significantly reduced the computational overhead.

$\text{GraphNOSE}_{\text{lin}}$ utilizing MolPCBA pre-training significantly outperforms the MPNN architecture\cite{sisson2025deep}, the best-performing model to date on this dataset (paired \(t\)-test, \(p=0.0083\) with only 281K parameters compared to the MPNN’s 2M parameters; other GPSE variants are numerically higher but untested (Table \ref{tab:mix}). Notably, linear models maintain highly competitive baseline scores for mixtures, with logistic regression utilizing Morgan fingerprint reaching an AUROC of 74.05\%. Overall, these results demonstrate that \textit{GraphNOSE} generalizes effectively to binary mixtures under a structure-disjoint split while remaining highly parameter efficient.



In summary, in this work we systematically resolved four open questions in computational olfaction. First, we demonstrated that augmenting graph transformers with learned GPSE representations significantly outperforms hand-crafted structural encodings (RWSE, LapPE). Furthermore, while the choice of GPSE pre-training corpus does not significantly alter predictive performance, MolPCBA was selected as the numerically optimal configuration and used throughout (Q1). Second, \textit{GraphNOSE} achieves robust performance in low-data regimes ($30\%\text{--}90\%$ training data) and strong generalization under extreme physico-chemical OOD shifts, outperforming baseline GNNs despite operating with a parameter-efficient footprint (Q2). Third, learned space neighborhood analysis and label distribution density estimations confirmed that the model's continuous latent space preserves hierarchical label clustering and perceptual similarities (Q3). Finally, by processing concatenated molecular graph pairs, \textit{GraphNOSE} successfully generalizes to binary odor mixtures, under a structure-disjoint split (Q4). Together, these empirical findings directly validate our initial hypotheses, confirming both that GPSE-derived spatial encodings deliver a superior inductive bias relative to hand-crafted encodings (hypothesis a) and that \textit{GraphNOSE} outperforms state-of-the-art MPNNs in structural generalization, low-data sample efficiency, and multi-label olfactory prediction (hypothesis b).

These findings substantially improve the accuracy–parameter trade-off in computational olfaction, balancing computational overhead with structural generalization and ecological validity. By integrating structural priors ($\text{GPSE}_{\text{MolPCBA}}$) with global attention in a parameter-efficient architecture, \textit{GraphNOSE} proves that superior predictive performance does not require massive computational overhead or static high-dimensional chemical fingerprints. By integrating single-molecule topology, binary mixture dynamics, and human perceptual space, this framework delivers a scalable and interpretable foundation for robust odor prediction.
Nevertheless, we observe sensitivity of the framework to variations in hyperparameter settings. As our model considers only the 2D representation, it assigns identical predictions to chiral enantiomers (mirror-image structures) that have distinct odors. Future work could address this limitation by incorporating explicit stereochemical features into node or edge representations. Furthermore, though we utilize exclusively the Performer based linear transformer, alternative sparse attention models such as BigBird, can be implemented in future work to assess potential gains in expressive capacity. Additionally, while highly computational, inherently non-linear deep learning models currently dominate predictive accuracy, there remains significant potential to improve classical linear models. Future work could focus on engineering richer, more expressive chemical descriptors that might allow these computationally lightweight, deterministic models to bridge the performance gap and potentially rival deep learning frameworks. Finally, the model was evaluated on curated GS-LF dataset. Future work is required to validate its scalability across larger, more diverse olfactory repositories to address open questions regarding robustness against label noise and the integration of concentration-dependent perceptual shifts that cannot be resolved using static binary datasets.

\section{Conclusions}
In this work, we have introduced  \textit{GraphNOSE}, a parameter-efficient graph transformer (GT) framework that bridges the complex non-linear gap between molecular structure and odor perception. We demonstrate that pre-trained structural and positional embeddings provide significantly richer inductive biases than do traditional explicitly constructed encodings. By combining chemical priors with a global attention mechanism, \textit{GraphNOSE} establishes a new benchmark by outperforming existing message passing architectures and quantum informed models while requiring significantly fewer parameters. Crucially, this framework not only offers a scalable solution for single molecules through the use of linear attention mechanisms but also successfully generalizes to the challenging task of prediction of binary odor mixtures and extrapolating to physico-chemically novel compounds. Ultimately, \textit{GraphNOSE} paves the way for future research in computational olfaction, serving as a foundational architecture for highly generalizable predictive modeling.

\subsection{Data and Software Availability}
The source code, Python scripts of charts, and other forms of analysis are available at the GitHub repository:  \url{https://github.com/CSIO-FPIL/GraphNOSE}. 

\subsection{Acknowledgments}
The authors acknowledge and are thankful to Council of Scientific and Industrial Research–Central Scientific Instruments Organisation (CSIR–CSIO), Sector 30, Chandigarh, for providing access to CSIR-4PI supercomputer. No external funding was involved in this project.

\subsection{Notes}
The authors declare no competing financial interests.
\newpage 


\clearpage 
\newpage

\begin{center}
    \LARGE \textbf{Supplementary Material\\ \textit{GraphNOSE}: A Graph Transformer in Olfaction }
\end{center}
\section{Contents}

\heading{1}{The Complex Relationship Between Molecular Structure and Odor}
\subentry{Figure S1: Structural similarity does not imply perceptual similarity}

\heading{2}{Dataset Analysis}
\subentry{Figure S2: Dataset insights}
\subentry{Table S1: Odor labels used in the curated dataset}

\heading{3}{Model Architecture}

\heading{4}{Loss Function}

\heading{5}{Baselines}

\heading{6}{Positional and Structural Encoding}

\heading{7}{Pre-training Dataset}
\subentry{Table S2: Summary of molecular datasets used in pre-training of GPSE}

\heading{8}{Hyper-parameter Search}
\subentry{Table S3: Hyper-parameters for \textit{GraphNOSE}}

\heading{9}{Training data efficiency and comparative GNN performance}
\subentry{Table S4: OpenPOM comparison in low-data regime}
\subentry{Table S5: $\text{LDM}_{\text{GCN}}$ comparison in low-data regime}
\subentry{Table S6: $\text{LDM}_{\text{DMPNN}}$ comparison in low-data regime}

\heading{10}{KDE Plot for True Labels}
\subentry{Figure S3: Analysis of odor space: KDE plot for true label}

\heading{11}{Explainability Analysis of Compounds}
\subentry{Figure S4: Explainability through integrated gradients for different odor descriptors}

\section{S1. The Complex Relationship Between Molecular Structure and Odor}

\begin{figure}[H]
    \centering
     \captionsetup{labelformat=empty}

     \includegraphics[width=\textwidth]{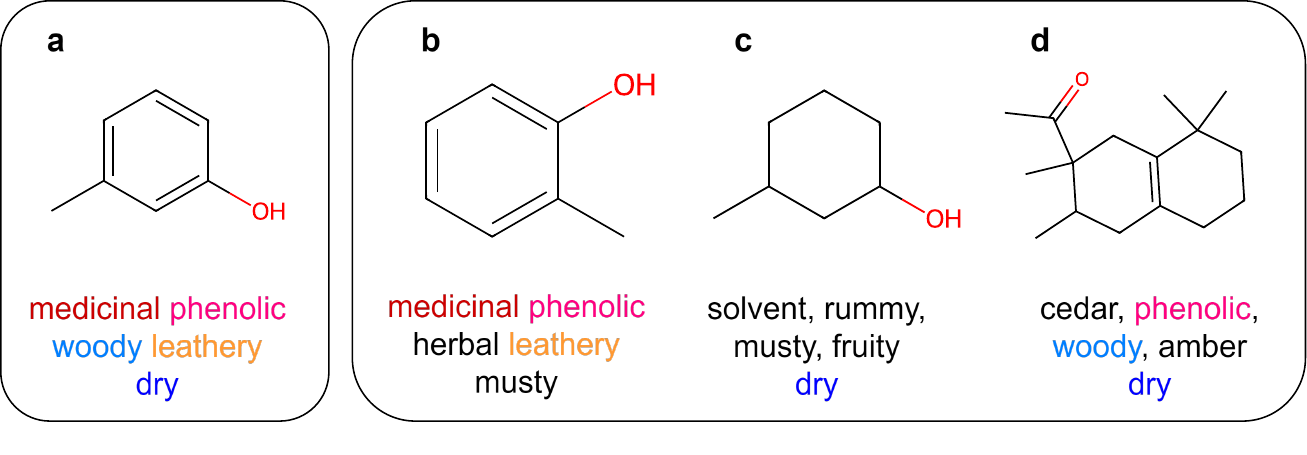}
    \caption{ Figure S1:\textbf{  Structural similarity does not imply perceptual similarity.} Perceptual odor similarity to a reference molecule (here, m-cresol, \textbf{a)} can be observed among some structurally related molecules (e.g., o-cresol, \textbf{b)} but not others (e.g., 3-methylcyclohexanol, \textbf{c)}, and structurally dissimilar molecules (e.g., a complex bicyclic ketone, \textbf{d)} can show odor similarity. Molecules sourced from the GF-LS dataset. \cite{luebke2019good, leffing}
 }
    \label{fig:strsim}
\end{figure}

\section{S2. Dataset Analysis}

 \begin{figure}[H]
    \centering
    \captionsetup{labelformat=empty}
    \begin{subfigure}[b]{\linewidth}
        \centering
        \includegraphics[width=\linewidth]{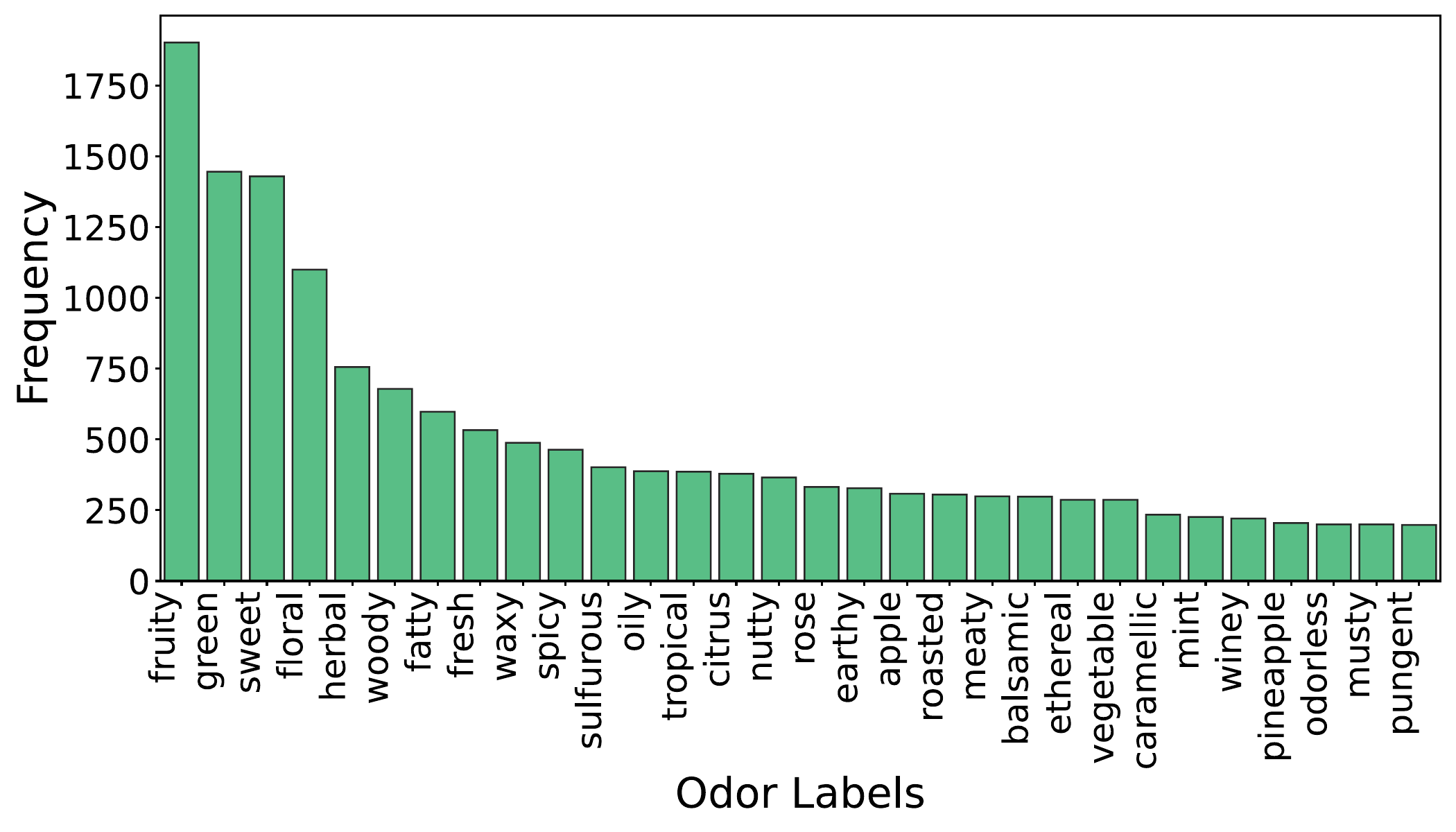}
        \caption{}
        \label{fig:modern_bar_chart}
    \end{subfigure}
    \hfill
    \begin{subfigure}[b]{\linewidth}
        \centering
        \includegraphics[width=\linewidth]{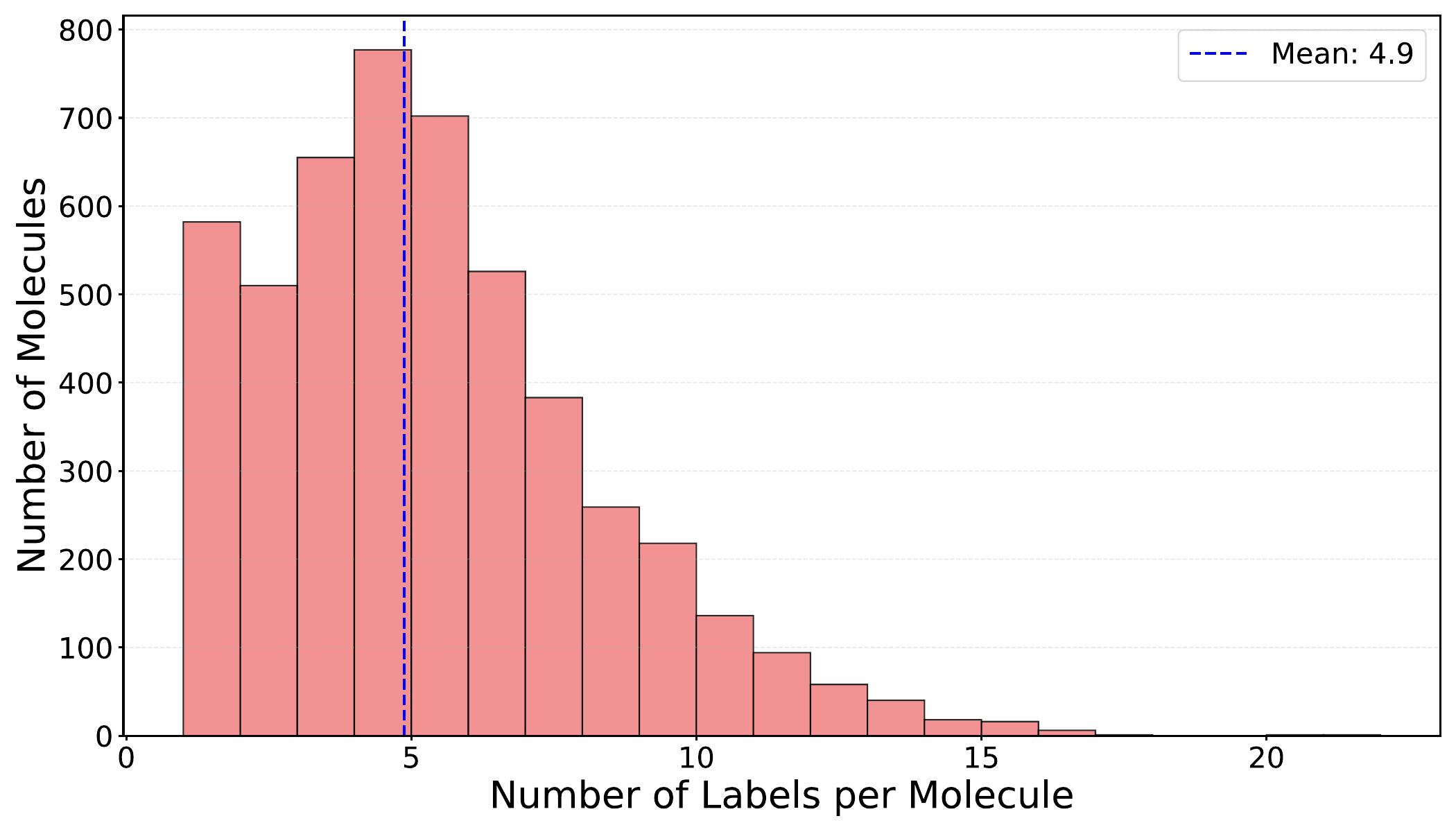}
        \caption{}
        \label{fig:labels_per_molecule}
    \end{subfigure}

    \caption{Figure S2: \textbf{Dataset insights: (a)} Frequency distribution of the top 30 odor labels among 138. (b) Number of labels per molecule in the GS-LF dataset\cite{aryan_amit_barsainyan_ritesh_kumar_pinaki_saha_michael_schmuker_2023} (total n = 4983).}
    \label{fig:odor_label_analysis}
\end{figure}

\begin{table}[H]
\centering
\small
\captionsetup{labelformat=empty}
\caption{Table S1: Odor labels used in the curated dataset.}
\begin{multicols}{5}
\begin{enumerate}[itemsep=0pt, parsep=0pt, topsep=0pt, leftmargin=*]
\item alcoholic
\item aldehydic
\item alliaceous
\item almond
\item amber
\item animal
\item anisic
\item apple
\item apricot
\item aromatic
\item balsamic
\item banana
\item beefy
\item bergamot
\item berry
\item bitter
\item black currant
\item brandy
\item burnt
\item buttery
\item cabbage
\item camphoreous
\item caramellic
\item cedar
\item celery
\item chamomile
\item cheesy
\item cherry
\item chocolate
\item cinnamon
\item citrus
\item clean
\item clove
\item cocoa
\item coconut
\item coffee
\item cognac
\item cooked
\item cooling
\item cortex
\item coumarinic
\item creamy
\item cucumber
\item dairy
\item dry
\item earthy
\item ethereal
\item fatty
\item fermented
\item fishy
\item floral
\item fresh
\item fruit skin
\item fruity
\item garlic
\item gassy
\item geranium
\item grape
\item grapefruit
\item grassy
\item green
\item hawthorn
\item hay
\item hazelnut
\item herbal
\item honey
\item hyacinth
\item jasmine
\item juicy
\item ketonic
\item lactonic
\item lavender
\item leafy
\item leathery
\item lemon
\item lily
\item malty
\item meaty
\item medicinal
\item melon
\item metallic
\item milky
\item mint
\item muguet
\item mushroom
\item musk
\item musty
\item natural
\item nutty
\item odorless
\item oily
\item onion
\item orange
\item orange\\ flower
\item orris
\item ozone
\item peach
\item pear
\item phenolic
\item pine
\item pineapple
\item plum
\item popcorn
\item potato
\item powdery
\item pungent
\item radish
\item raspberry
\item ripe
\item roasted
\item rose
\item rummy
\item sandalwood
\item savory
\item sharp
\item smoky
\item soapy
\item solvent
\item sour
\item spicy
\item strawberry
\item sulfurous
\item sweaty
\item sweet
\item tea
\item terpenic
\item tobacco
\item tomato
\item tropical
\item vanilla
\item vegetable
\item vetiver
\item violet
\item warm
\item waxy
\item weedy
\item winey
\item woody
\end{enumerate}
\end{multicols}
\label{tab:odor_labels}
\end{table}

\section{S3. Model Architecture}
\label{sec:Model_Architecture}
GraphNOSE model has been built along the lines of the GraphGPS architecture\cite{rampavsek2022recipe} using \textit{PyTorch Geometric} \cite{fey2019fast}. The important difference in our implementation is that we don't update the edge attributes. Each node in the input graph has an attribute and positional and structural encoding (PSE) associated with it. Each edge in the graph is associated with a edge attribute. The PSE is projected using a multi-layer perceptron (MLP) to the projection dimension. The result is concatenated to the original node features and processed through a linear layer that  maps it to a hidden dimension. Each edge attribute is also mapped to the \emph{hidden dimension} using a linear layer. This graph with updated node and edge features is passed through 4 layers of GPS (General, Powerful and Scalable architecture) \cite{rampavsek2022recipe} layers. A readout layer of global mean pooling is used on the final node embeddings. This pooled embedding is finally passed through an MLP for the final multi-label classification. 

During the hyperparameter search for GraphNOSE, the global attention type in GPS layer was fixed to Performer \cite{choromanski2020rethinking}. The search results showed that GPSE pre-trained on MoLPCBA was the optimal PSE setting. This exact architecture was used while evaluating other PSEs described in the main text of this paper.

More information about the hyper-parameter tuning of GraphNOSE can be found in Section \hyperref[sec:Hyperparameter_search]{S8}. The exact values of the above-mentioned hyperparameters can be found in Table \hyperref[tab:graphnose_hparams]{S3} in that section..

\setcounter{table}{0}
\setcounter{figure}{0}
\renewcommand{\thetable}{\thesection.\arabic{table}}
\renewcommand{\thefigure}{\thesection.\arabic{figure}}
\section{S4. Loss Function}
\label{sec:Loss_Function}
\setcounter{table}{0}
\setcounter{figure}{0}
\renewcommand{\thetable}{\thesection.\arabic{table}}
\renewcommand{\thefigure}{\thesection.\arabic{figure}}

Whenever we work on a multi-label odor dataset, the distribution of the label sets and degree of data imbalance play vital roles in model generalizability and predictive accuracy. In highly imbalanced datasets, models often exhibit high micro-averaged ROC scores but suffer from a low macro-averaged ROCs as the model fails to generalize to under-represented descriptors. Though various measures help to deal with this issue, such as the co-occurrence level measure and the imbalance measures, this study employs imbalance ratio per label \cite{tarekegn2021review} to analyze the label-wise distribution.

Formally, let $\mathcal{D}$ be a multi-label dataset with a label set $L$, where $Y_i$ represents the label set for the $i$\textsuperscript{th} instance. The \textit{IRLbl} for a specific label $\lambda \in L$ is defined as the ratio of the frequency of the most common label to the frequency of label $\lambda$.\begin{equation}IRLbl(\lambda) = \frac{\max_{\lambda' \in L} \sum_{i=1}^{m} h(\lambda', Y_i)}{\sum_{i=1}^{m} h(\lambda, 
Y_i)}\end{equation}
where the indicator function $h(\lambda, Y_i)$ is defined as:\begin{equation}
h(\lambda, Y_i) =
\begin{cases}
1 & \text{if } \lambda \in Y_i \\
0 & \text{if } \lambda \notin Y_i
\end{cases}
\end{equation}
Under this metric, the most frequent label receives an IRLbl of 1; higher values signify a greater degree of imbalance for that specific descriptor. Motivated by POM \cite{lee2023principal}, we implemented a loss which is weighted by of $log(1+ class\_imbalance_ratio)$

\section{S5. Baselines}

\setcounter{table}{0}
\setcounter{figure}{0}
\renewcommand{\thetable}{\thesection.\arabic{table}}
\renewcommand{\thefigure}{\thesection.\arabic{figure}}
\label{sec:GNN_Baselines_appendix}
\subsection{OpenPOM}
OpenPOM \cite{OpenPOM} is a publicly available version of the Principal Odor Map paper \cite{lee2023principal}and integrates with Deepchem \cite{Olfaction}. It uses a mono molecular odor dataset consisting of 4983 molecules \cite{aryan_amit_barsainyan_ritesh_kumar_pinaki_saha_michael_schmuker_2023} curated by merging samples from two expert sources: GoodScents\cite{luebke2019good} and Leffingwell PMP 2001\cite{leffing}. OpenPOM addresses the problem of multi-label classification for the 138 odor labels of molecules represented as SMILES strings. The model predicts the probability for each of the 138 possible odor descriptors corresponding to a given molecule by utilizing a message-passing neural network (MPNN), which maps chemical structures to perceptually meaningful latent space. The MPNN sends messages between nodes to understand the local chemical neighborhood and then pools the node features into a global molecular embedding. The embedding so obtained is fed into a feed-forward network with a sigmoid activation layer at the end to output probabilities for the 138 odor labels. OpenPOM uses a custom featurizer, with 134 node features, which include hydrogen count, valence, formal charge, degree, hybridization, atomic number; and 6 edge features, which include aromaticity, degree, and whether it is in a ring. All these features are one-hot encoded (i.e., represented as binary vectors where a single active bit indicates the presence of a specific feature while all other elements are zero). OpenPOM provides a standardized benchmark that bridges chemical structure to odor perception.
\subsection{Models Based on Atoms in Molecule Localization and Delocalization Matrices (AIMLDM)
}

This work by Saha \textit{et al}\cite{D5DD00224A} is a paradigm shift where the quantum mechanical features are employed as the node and edge features along with the chemically grounded featurizers. The model is trained on 4872  molecules, a subset of the curated OpenPOM dataset \cite{aryan_amit_barsainyan_ritesh_kumar_pinaki_saha_michael_schmuker_2023}, using a summed binary cross-entropy loss weighted by imbalance ratio per label (IRLbl)\cite{tarekegn2021review} for training.
\subsubsection{\texorpdfstring{$\text{LDM}_{\text{DMPNN}}$}{LDM-DMPNN}}
In this methodology the dataset goes through two featurization pipelines. The first is constructed through the quantum features using the AIMLDM  approach. In this approach localization indices (diagonal terms) and delocalization indices (off diagonal terms) are taken as node features and edge features, respectively. The second featurization pipeline employs DMPNN Featurizer,12 which captures 133 node and 14 edge features. Consequently, the model incorporates the chemical features, bond connectivity, and atom types, as well as the quantum features, representing electron density distribution. Of the two MPNNs, the one with chemical features is processed by utilizing the Set2Set readout, and the other with quantum features is processed through global sum pooling to obtain a global molecular embedding for each of the pipelines. The embeddings from both branches are concatenated and fed into a feed-forward neural network. The training process employs the Adam optimizer with exponential learning rate decay and uses the AUROC metric to evaluate the performance.
\subsubsection{\texorpdfstring{$\text{LDM}_{\text{GCN}}$}{LDM-GCN}}
This methodology also utilizes the same graph neural network (GNN) architecture as LDMDMPNN. The architecture is same, except that it employs the MolGraph-ConvFeaturizer \cite{kearnes2016molecular} featurizer and has 30 node and 11 edge features. As reported in the main text, LDMGCN achieves a better performance than the odor prediction model that employs the LDM matrix coupled with MPNN.
\subsection{MPNN model: Aroma Blend Prediction}
\label{mix_model}
The MPNN model proposed by Sisson \textit{et al} \cite{sisson2025deep} utilizes exactly the same node and edge attributes employed by OpenPOM (134 node and 6 edge attributes)8. Molecular structures for pairs of binary odor mixture molecules were grouped into a unified two-component graph. The aggregation layer treats all the atoms as if they belong to the same molecule and generates a final permutation invariant embedding. The model aggregates molecular embeddings using a Set2Set readout function. Training is performed using a summed binary cross-entropy loss weighted by the imbalance ratio per label (IRLbl) to address class imbalance. To evaluate the performance on the binary mixture molecules, it applies a “graph carving” technique that creates a 50:50 train/test split by maximizing the usable samples and minimizing the distribution shift. It evaluates the model on 74 shared odor labels and strictly prevents data leakage by ensuring no individual molecule appears in both partitions. Our work also uses the same split as proposed in the Sisson \textit{et al.} model.
\section{S6. Positional and structural encoding }

\label{sec:Positional_and_structural_encoding}
We employ two complementary types of node-level encodings to capture the local and global topology of the molecular graphs: random-walk structural encodings (RWSE) and Laplacian positional encodings (LapPE).

\subsection{Random-walk structural encodings (RWSE)}
RWSE characterizes the local neighborhood of a node $i$ by observing the probability that a random walker returns to the starting node after $k$ steps. For a graph with adjacency matrix $A$ and degree matrix $D$, the random walk transition matrix is defined as $P = D^{-1}A$. The RWSE for node $i$ is a vector defined by
\begin{equation}
    \text{RWSE}(i) = [P^1_{ii}, P^2_{ii}, \dots, P^k_{ii}]
\end{equation}
This encoding identifies local structural motifs, such as aromatic rings and functional groups, by capturing the density of local cycles. 
For our experiment  $k=16$.

\subsection{Laplacian Eigenvector Positional Encodings (LapPE)}

The Laplacian eigenvector positional encodings (LapPE)  are derived from the eigen decomposition of the graph Laplacian matrix $L$.\cite{canturk2023graph} For a simple undirected and unweighted graph $G = (V, E)$, the Laplacian is defined as
\begin{equation}
    L = D - M \label{eq:laplacian}
\end{equation}
where $D \in \mathbb{N}^{n \times n}$ is the diagonal degree matrix with $D_{ii} = \text{deg}(v_i)$, and $M \in \{0, 1\}^{n \times n}$ is the symmetric adjacency matrix. 

The real symmetric matrix $L$ admits a full eigen decomposition:
\begin{equation}
    L = U\Lambda U^\top \label{eq:eigendecomp}
\end{equation}
where $\Lambda_{ii} = \lambda_i$ and $U_{[:,i]} = u_i$ represent the $i$-th eigenvalue and eigenvector, respectively. Following standard convention, these are indexed such that $0 = \lambda_1 \le \lambda_2 \le \dots \le \lambda_n$. 

We use the 8 non-trivial eigenvectors of the graph Laplacian as the positional encoding.

\newpage
\section{S7. Pre-training datasets }
\label{pre_trained}
\
\label{sec:Pre-training_datasets}

\begin{table}[ht]
\centering
\captionsetup{labelformat=empty}
\caption{Table S2: Summary of molecular datasets used in pretraining of GPSE.}
\label{tab:datasets}
\begin{tabular}{|l|l|l|r|}
\hline
\textbf{Dataset} & \textbf{Primary Source} & \textbf{License} & \textbf{Unique Graphs} \\
\hline
MolPCBA  & \makecell[l]{MoleculeNet \\ \cite{hu2020open}} 
         & MIT 
         & 323{,}555 \\
\hline
ZINC     & \makecell[l]{ZINC Database \\ \cite{gomez2018automatic}} 
         & Apache 2.0 License
         & 249{,}455 \\
\hline
GEOM     & \makecell[l]{Axelrod \& Gomez-Bombarelli \\ \cite{axelrod2022geom}} 
         & CC0 1.0 
         & 169{,}925 \\
\hline
ChEMBL   & \makecell[l]{Gaulton et al. \\ \cite{gaulton2012chembl}} 
         & CC BY-SA 3.0 
         & 970{,}963 \\
\hline
PCQM4Mv2 & \makecell[l]{PubChemQC \\ \cite{hu2021ogb}} 
         & CC BY 4.0 
         & 273{,}920 \\
\hline
\end{tabular}
\end{table}

\section{S8. Hyperparameter Search}
\label{sec:Hyperparameter_search}
For random forest (RF) baseline methods, we tuned the number of estimators, maximum depths of each tree, minimum number of samples required to split an internal node, minimum number of samples required to be a leaf node, and number of features to consider when looking for the best split. We also tune the fingerprinting parameters such as radius and length. 

For the  k-nearest-neighbor (KNN) models, we tuned the value of \textit{k} for both GraphNOSE and Mordred embeddings setups. For the Mordred setup, we additionally tuned the radius and length of the fingerprint. The KNN predictions are distance weighted. 

For the LLM baselines, we derive the embeddings of SMILES using the pre-trained weights of MoLFormer\cite{ross2022large} and ChemBERTa\cite{chithrananda2020chemberta}. We tune the hidden dimension of the MLP classifier along with the other common hyperparameters such as batch size, learning rate and weight decay.

\begin{table}[H]
\centering
\captionsetup{labelformat=empty}
\caption{Table S3: Hyperparameters for GraphNOSE}
\label{tab:graphnose_hparams}
\small
\begin{tabular}{|l|c|c|}
\hline
\textbf{Hyperparameter} & \textbf{Search Space} & \textbf{Optimal Value} \\
\hline

\multicolumn{3}{|l|}{\textit{Architecture Parameters}} \\
\hline
Hidden dimension & \{64, 128, 256, 288, 352, 512\} & 64 \\
\hline
Classifier dimension & \{128, 256, 512, 1024\} & 512 \\
\hline
Heads & \{1, 2, 4\} & 1 \\
\hline
No. of GPS layers & \{4, 5, 6, 7\} & 4 \\
\hline
Dropout & [0.0, 0.5] & 0.440 \\
\hline
Local GNN & \{gatv2, resgate, gine\} & gine \\
\hline
Activation function & \{relu, gelu, elu, silu\} & relu \\
\hline

\multicolumn{3}{|l|}{\textit{Positional Encoding Parameters}} \\
\hline
PSE & \{random\_walk\_pe, laplacian\_pe, & gpse\_molpcba \\
 & gpse\_zinc, gpse\_chembl, & \\
 & gpse\_molpcba, gpse\_pcqm4mv2, & \\
 & gpse\_geom\} & \\
\hline
Projection dimension & \{4, 8, 16, 32\} & 16 \\
\hline

\multicolumn{3}{|l|}{\textit{Training Parameters}} \\
\hline
Learning rate & [1e-5, 1e-2], log scale & 6.82e-4 \\
\hline
Weight decay & [1e-6, 1e-3], log scale & 9.43e-5 \\
\hline
Batch size & \{16, 32, 64, 128\} & 16 \\
\hline
Minimum learning rate & [1e-7, 1e-5], log scale & 1.98e-7 \\
\hline
Optimizer & \{adam, adamw, sgd\} & adam \\
\hline

\end{tabular}
\end{table}


\newpage
\section{S9. Training data efficiency and comparative GNN performance}
\label{sec:training_search}

\begin{table}[H]
\centering
\captionsetup{labelformat=empty}
\caption{Table S4: AUROC on the held-out test set across five random seeds and paired $t$-test $p$-values comparing OpenPOM against $\text{GraphNOSE}_{\text{lin}}$ and $\text{GraphNOSE}_{\text{trans}}$ across training data proportions (30\% -- 90\%). Significance levels ($^*$ denote paired $t$-test comparison): $^{***}p < 0.001$, $^{**}p < 0.01$, and $^{*}p < 0.05$.}
\label{tab:openpom_fraction_stats}
\resizebox{\textwidth}{!}{%
\begin{tabular}{|l|c|c|c|c|c|c|c|}
\hline
\textbf{Sampled Train} & \textbf{Seed 1} & \textbf{Seed 10} & \textbf{Seed 100} & \textbf{Seed 21} & \textbf{Seed 42} & \makecell{\textbf{$p$-val} \\ \textbf{(vs. $\text{GraphNOSE}_{\text{lin}}$)}} & \makecell{\textbf{$p$-val} \\ \textbf{(vs. $\text{GraphNOSE}_{\text{trans}}$)}} \\ \hline
30\% & 82.32 & 81.88 & 82.62 & 80.55 & 82.40 & $1.51 \times 10^{-3\,**}$ & $1.43 \times 10^{-3\,**}$ \\ \hline
40\% & 83.91 & 83.52 & 84.12 & 84.01 & 84.06 & $4.39 \times 10^{-4\,***}$ & $1.55 \times 10^{-4\,***}$ \\ \hline
50\% & 84.35 & 84.35 & 84.01 & 84.95 & 84.56 & $2.95 \times 10^{-4\,***}$ & $1.06 \times 10^{-4\,***}$ \\ \hline
60\% & 85.01 & 85.54 & 85.38 & 85.15 & 85.20 & $3.00 \times 10^{-5\,***}$ & $2.80 \times 10^{-5\,***}$ \\ \hline
70\% & 86.72 & 86.96 & 86.58 & 86.90 & 86.51 & $7.59 \times 10^{-4\,***}$ & $1.10 \times 10^{-4\,***}$ \\ \hline
80\% & 87.03 & 87.07 & 86.92 & 87.48 & 86.65 & $2.87 \times 10^{-4\,***}$ & $5.70 \times 10^{-4\,***}$ \\ \hline
90\% & 87.50 & 87.03 & 86.90 & 87.27 & 86.48 & $9.18 \times 10^{-4\,***}$ & $1.87 \times 10^{-4\,***}$ \\ \hline
\end{tabular}%
}
\end{table}

\begin{table}[H]
\centering
\captionsetup{labelformat=empty}
\caption{Table S5: AUROC on the held-out test set across five random seeds and paired $t$-test $p$-values comparing $\text{LDM}_{\text{GCN}}$ against $\text{GraphNOSE}_{\text{lin}}$ and $\text{GraphNOSE}_{\text{trans}}$ across training data proportions (30\% -- 90\%). Significance levels ($^*$ denote paired $t$-test comparison): $^{***}p < 0.001$, $^{**}p < 0.01$, and $^{*}p < 0.05$.}
\label{tab:ldm_gcn_fraction_stats}
\resizebox{\textwidth}{!}{%
\begin{tabular}{|l|c|c|c|c|c|c|c|}
\hline
\textbf{Sampled Train} & \textbf{Seed 1} & \textbf{Seed 10} & \textbf{Seed 100} & \textbf{Seed 21} & \textbf{Seed 42} & \makecell{\textbf{$p$-val} \\ \textbf{(vs. $\text{GraphNOSE}_{\text{lin}}$)}} & \makecell{\textbf{$p$-val} \\ \textbf{(vs. $\text{GraphNOSE}_{\text{trans}}$)}} \\ \hline
30\% & 82.71 & 77.31 & 80.97 & 82.09 & 81.56 & $1.50 \times 10^{-2\,*}$ & $1.29 \times 10^{-2\,*}$ \\ \hline
40\% & 83.57 & 83.45 & 82.74 & 83.24 & 83.85 & $5.10 \times 10^{-5\,***}$ & $3.60 \times 10^{-5\,***}$ \\ \hline
50\% & 83.80 & 84.80 & 83.87 & 84.97 & 84.72 & $2.79 \times 10^{-4\,***}$ & $2.11 \times 10^{-4\,***}$ \\ \hline
60\% & 84.68 & 84.91 & 83.83 & 84.43 & 85.10 & $1.38 \times 10^{-4\,***}$ & $1.80 \times 10^{-4\,***}$ \\ \hline
70\% & 85.01 & 85.30 & 85.85 & 85.31 & 85.56 & $1.60 \times 10^{-5\,***}$ & $2.30 \times 10^{-5\,***}$ \\ \hline
80\% & 86.16 & 86.18 & 86.13 & 86.08 & 86.56 & $1.00 \times 10^{-5\,***}$ & $2.31 \times 10^{-7\,***}$ \\ \hline
90\% & 86.32 & 86.49 & 86.50 & 86.77 & 86.68 & $2.10 \times 10^{-5\,***}$ & $6.50 \times 10^{-5\,***}$ \\ \hline
\end{tabular}%
}
\end{table}

\begin{table}[H]
\centering
\captionsetup{labelformat=empty}
\caption{Table S6: AUROC on the held-out test set across five random seeds and paired $t$-test $p$-values comparing $\text{LDM}_{\text{DMPNN}}$ against $\text{GraphNOSE}_{\text{lin}}$ and $\text{GraphNOSE}_{\text{trans}}$ across training data proportions (30\% -- 90\%). Significance levels ($^*$ denote paired $t$-test comparison): $^{***}p < 0.001$, $^{**}p < 0.01$, and $^{*}p < 0.05$.}
\label{tab:ldm_dmpnn_fraction_stats}
\resizebox{\textwidth}{!}{%
\begin{tabular}{|l|c|c|c|c|c|c|c|}
\hline
\textbf{Sampled Train} & \textbf{Seed 1} & \textbf{Seed 10} & \textbf{Seed 100} & \textbf{Seed 21} & \textbf{Seed 42} & \makecell{\textbf{$p$-val} \\ \textbf{(vs. $\text{GraphNOSE}_{\text{lin}}$)}} & \makecell{\textbf{$p$-val} \\ \textbf{(vs. $\text{GraphNOSE}_{\text{trans}}$)}} \\ \hline
30\% & 81.76 & 82.40 & 82.10 & 81.81 & 81.67 & $3.00 \times 10^{-6\,***}$ & $1.66 \times 10^{-6\,***}$ \\ \hline
40\% & 83.96 & 83.90 & 83.39 & 81.82 & 83.69 & $1.70 \times 10^{-3\,**}$ & $2.55 \times 10^{-3\,**}$ \\ \hline
50\% & 83.87 & 83.30 & 82.95 & 83.58 & 84.03 & $3.71 \times 10^{-4\,***}$ & $1.62 \times 10^{-4\,***}$ \\ \hline
60\% & 85.49 & 84.76 & 84.70 & 85.02 & 84.86 & $5.20 \times 10^{-5\,***}$ & $3.40 \times 10^{-5\,***}$ \\ \hline
70\% & 85.88 & 85.63 & 85.46 & 85.74 & 83.94 & $1.46 \times 10^{-3\,**}$ & $1.06 \times 10^{-3\,**}$ \\ \hline
80\% & 85.94 & 86.08 & 86.32 & 85.70 & 86.12 & $7.50 \times 10^{-5\,***}$ & $1.90 \times 10^{-5\,***}$ \\ \hline
90\% & 86.27 & 86.36 & 86.64 & 86.42 & 86.46 & $9.00 \times 10^{-6\,***}$ & $1.50 \times 10^{-5\,***}$ \\ \hline
\end{tabular}%
}
\end{table}
\section{S10. KDE Plot for True Labels}

\begin{figure}[H]
    \centering
    \captionsetup{labelformat=empty}
     \includegraphics[scale=0.05]{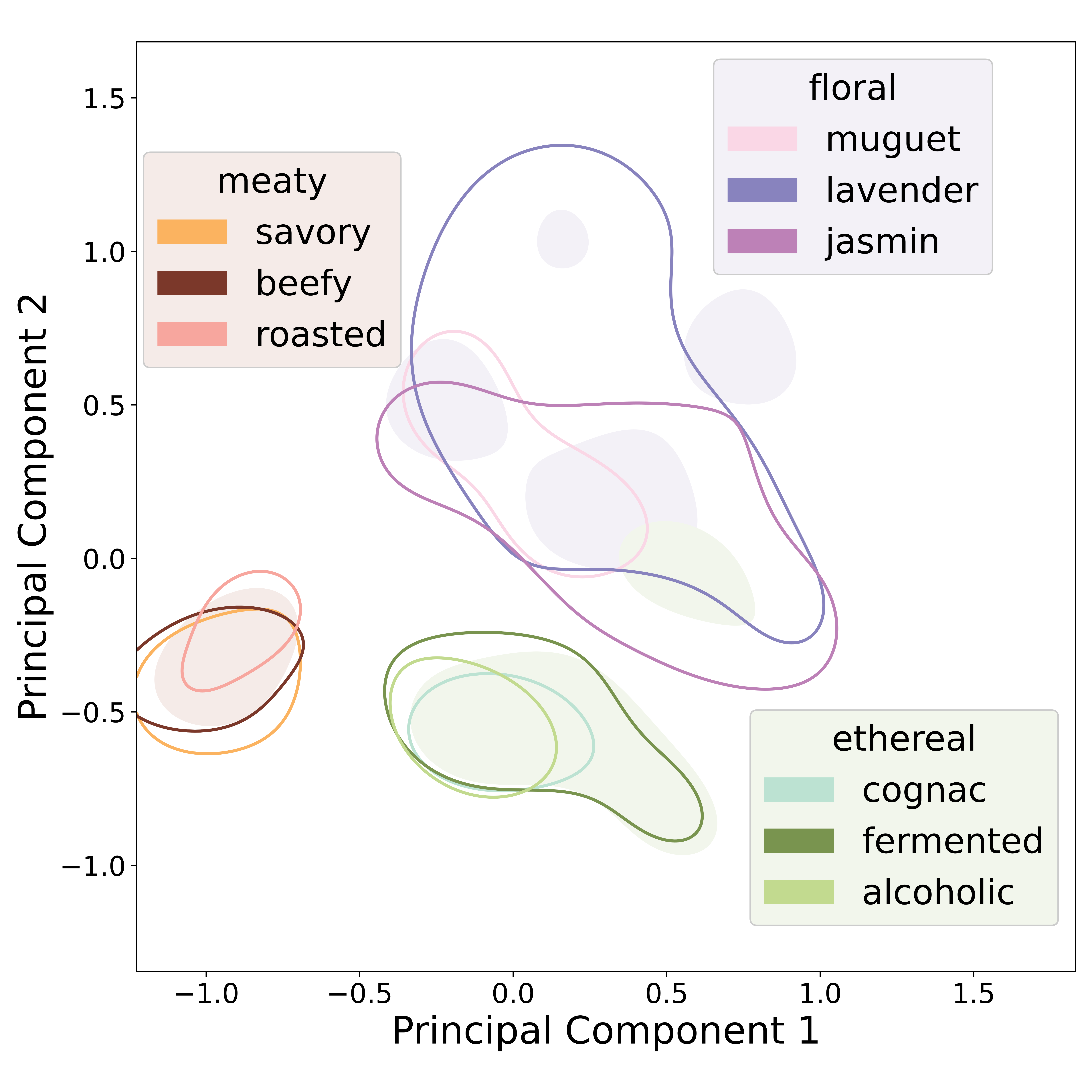}
    \caption{Figure S3: Analysis of odor space: KDE plot for true label. Shaded and contoured areas are kernel density estimates of the
distribution of labeled data. Colors denote odor categories, with shaded areas for broad classes and contours for specific descriptors (e.g., lavender, roasted).}
    \label{fig:kde_svg}
\end{figure}

\section{S11. Explainability Analysis of Compounds}

\label{sec:Explainability_analysis}
\begin{figure}[H]
    \centering
\captionsetup{labelformat=empty}

    \begin{subfigure}{\textwidth}
        \centering
        \begin{minipage}[b]{0.32\textwidth}
            \includegraphics[width=\textwidth]{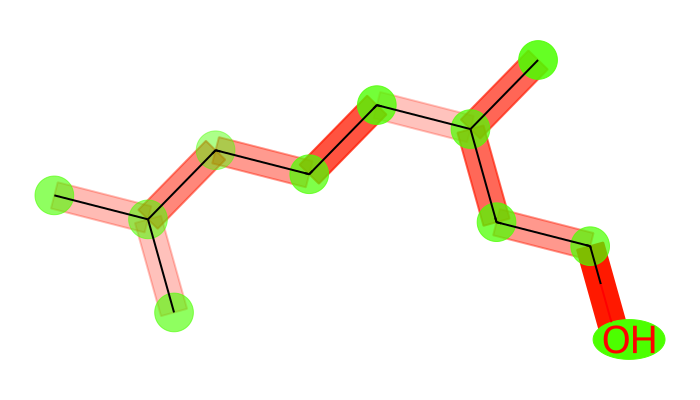}
        \end{minipage}
        \hfill
        \begin{minipage}[b]{0.32\textwidth}
            \includegraphics[width=\textwidth]{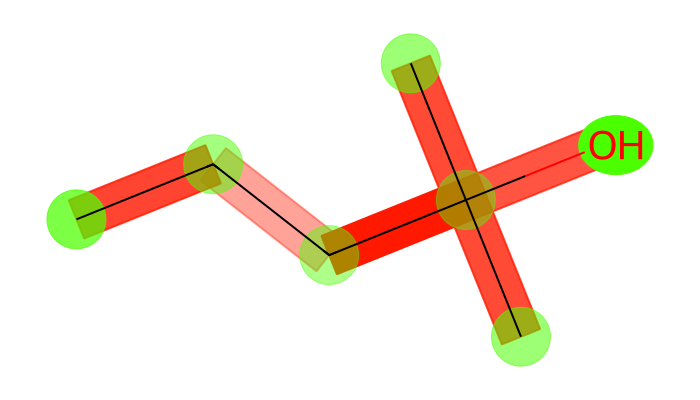}
        \end{minipage}
        \hfill
        \begin{minipage}[b]{0.32\textwidth}
            \includegraphics[width=\textwidth]{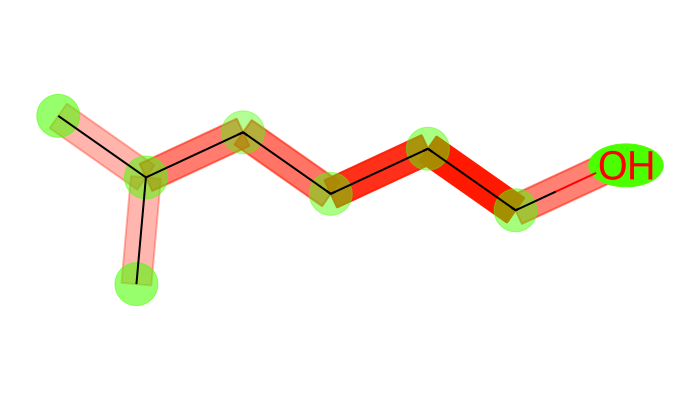}
        \end{minipage}
        \caption{}
        \label{fig:alcohol_row}
    \end{subfigure}

    \medskip

    \begin{subfigure}{\textwidth}
        \centering
        \begin{minipage}[b]{0.24\textwidth}
            \includegraphics[width=\textwidth]{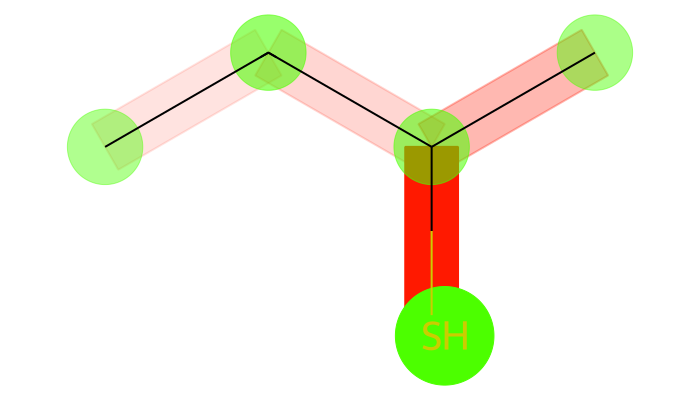}
        \end{minipage}
        \hfill
        \begin{minipage}[b]{0.24\textwidth}
            \includegraphics[width=\textwidth]{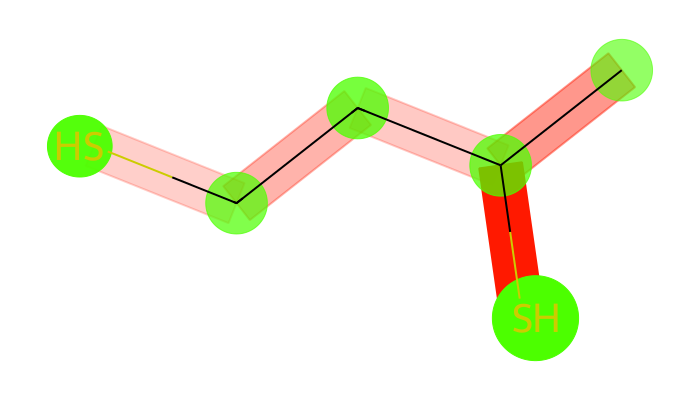}
        \end{minipage}
        \hfill
        \begin{minipage}[b]{0.24\textwidth}
            \includegraphics[width=\textwidth]{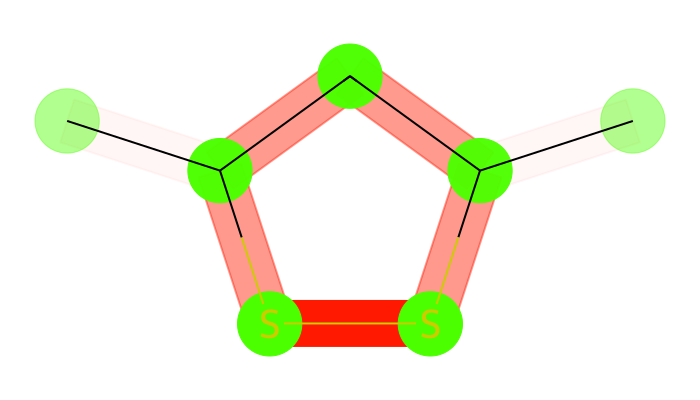}
        \end{minipage}
        \hfill
        \begin{minipage}[b]{0.24\textwidth}
            \includegraphics[width=\textwidth]{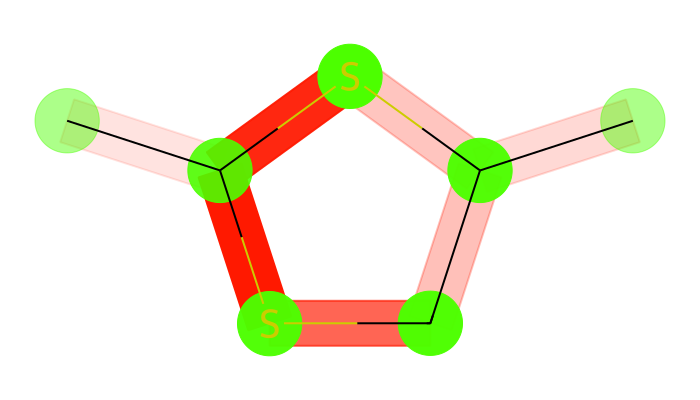}
        \end{minipage}
        \caption{}
        \label{fig:sulphur_row}
    \end{subfigure}

    \medskip

    \begin{subfigure}{\textwidth}
        \centering
        \begin{minipage}[b]{0.24\textwidth}
            \includegraphics[width=\textwidth]{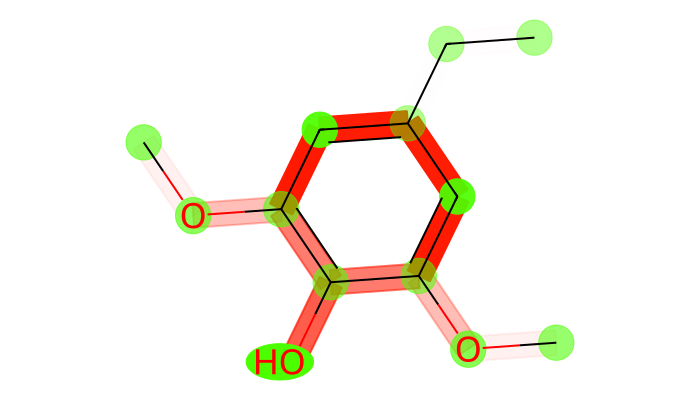}
        \end{minipage}
        \hfill
        \begin{minipage}[b]{0.24\textwidth}
            \includegraphics[width=\textwidth]{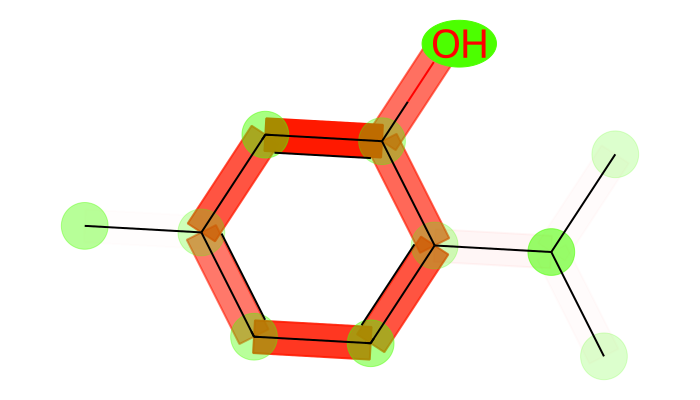}
        \end{minipage}
        \hfill
        \begin{minipage}[b]{0.24\textwidth}
            \includegraphics[width=\textwidth]{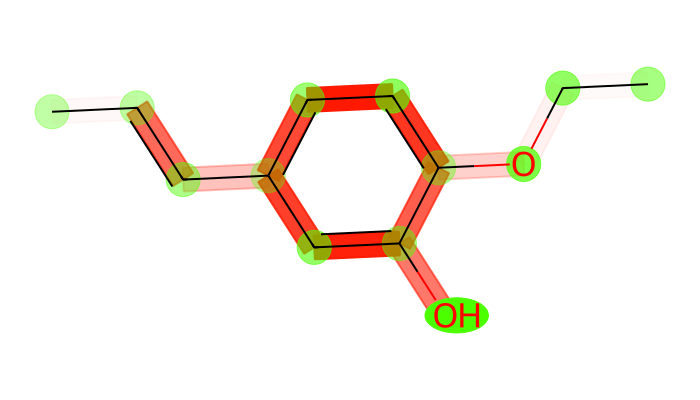}
        \end{minipage}
        \hfill
        \begin{minipage}[b]{0.24\textwidth}
            \includegraphics[width=\textwidth]{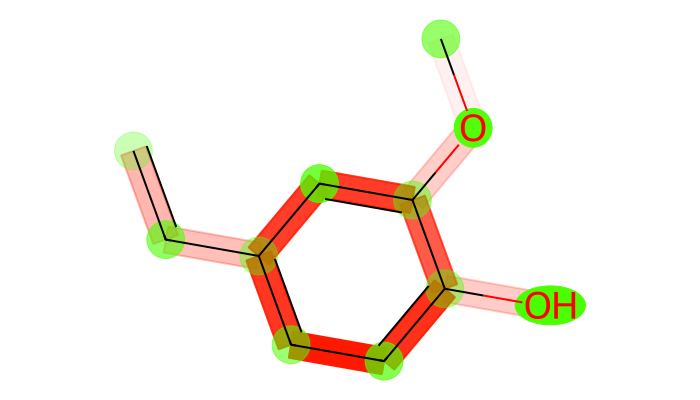}
        \end{minipage}
        \caption{}
        \label{fig:phenolic_row}
    \end{subfigure}

    \medskip

    \begin{subfigure}{\textwidth}
        \centering
        \begin{minipage}[b]{0.24\textwidth}
            \includegraphics[width=\textwidth]{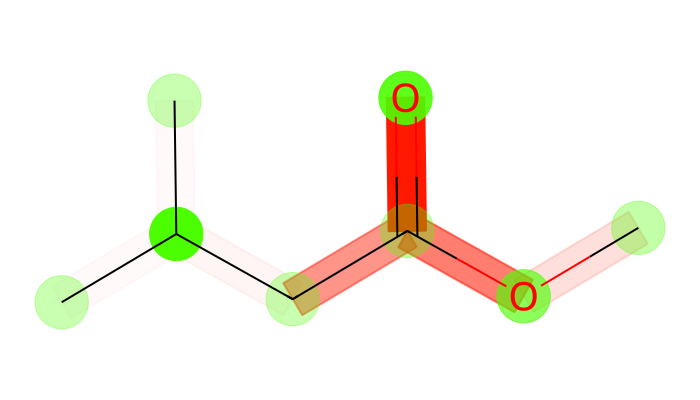}
        \end{minipage}
        \hfill
        \begin{minipage}[b]{0.24\textwidth}
            \includegraphics[width=\textwidth]{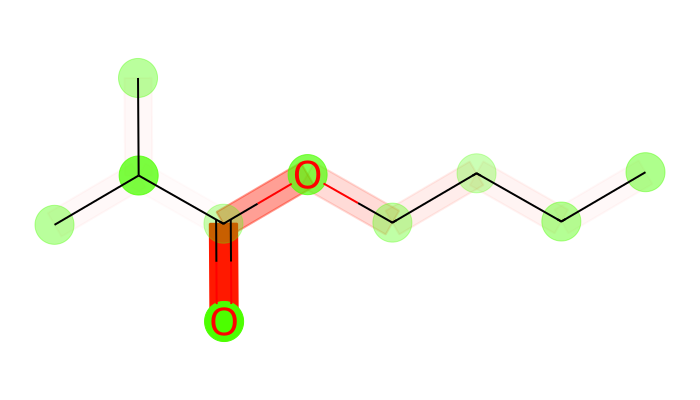}
        \end{minipage}
        \hfill
        \begin{minipage}[b]{0.24\textwidth}
            \includegraphics[width=\textwidth]{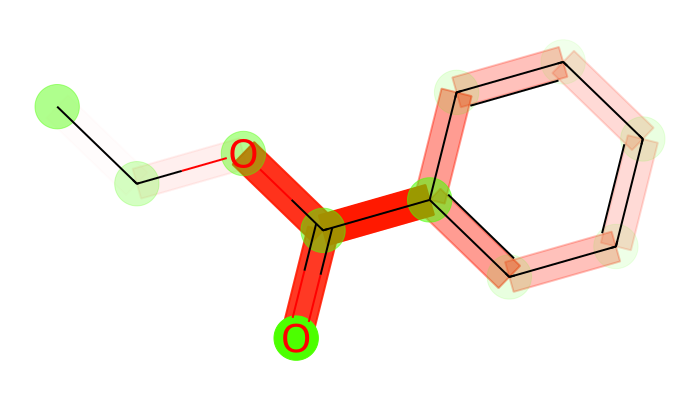}
        \end{minipage}
        \hfill
        \begin{minipage}[b]{0.24\textwidth}
            \includegraphics[width=\textwidth]{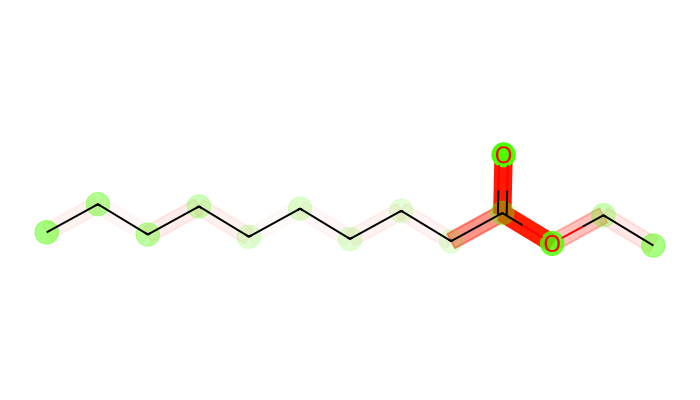}
        \end{minipage}
        \caption{}
        \label{fig:fruity_row}
    \end{subfigure}

    \medskip

    \begin{subfigure}{\textwidth}
        \centering
        \begin{minipage}[b]{0.24\textwidth}
            \includegraphics[width=\textwidth]{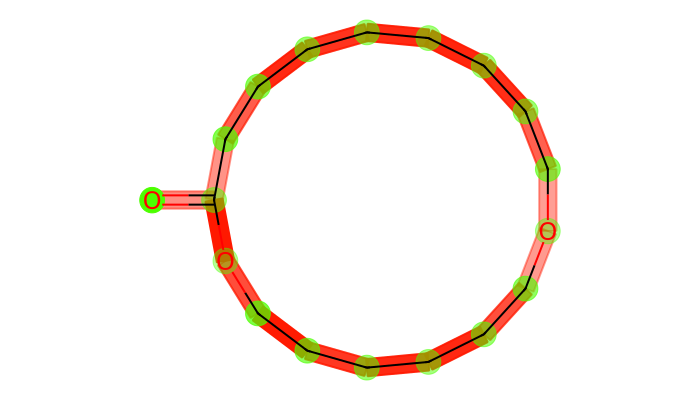}
        \end{minipage}
        \hfill
        \begin{minipage}[b]{0.24\textwidth}
            \includegraphics[width=\textwidth]{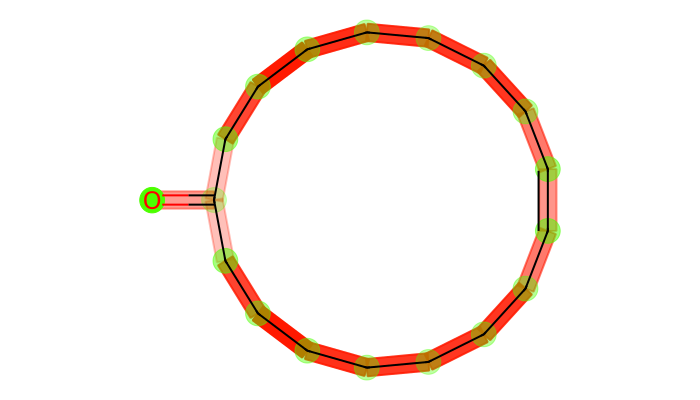}
        \end{minipage}
        \hfill
        \begin{minipage}[b]{0.24\textwidth}
            \includegraphics[width=\textwidth]{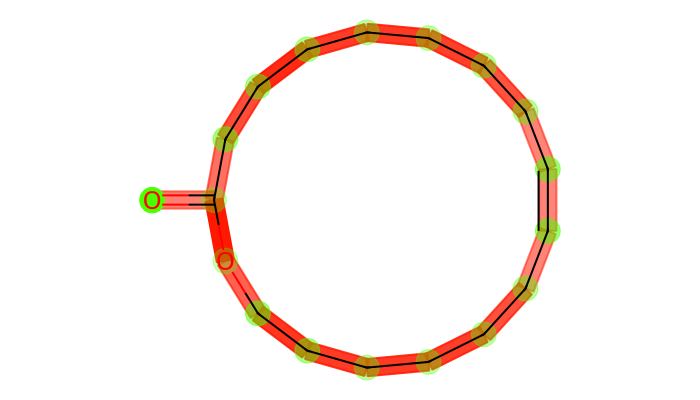}
        \end{minipage}
        \hfill
        \begin{minipage}[b]{0.24\textwidth}
            \includegraphics[width=\textwidth]{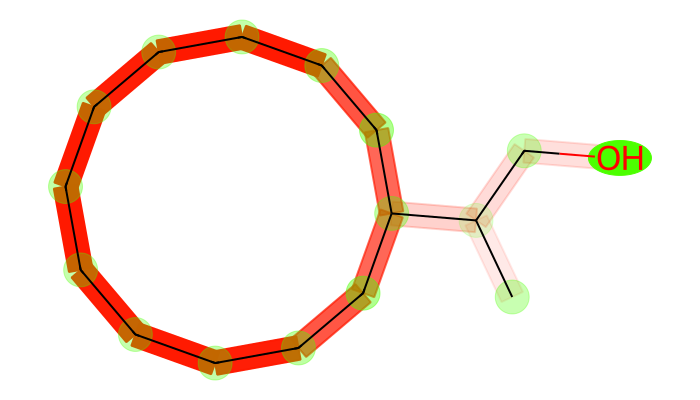}
        \end{minipage}
        \caption{}
        \label{fig:musk_row}
    \end{subfigure}

    \medskip

    \begin{subfigure}{\textwidth}
        \centering
        \begin{minipage}[b]{0.24\textwidth}
            \includegraphics[width=\textwidth]{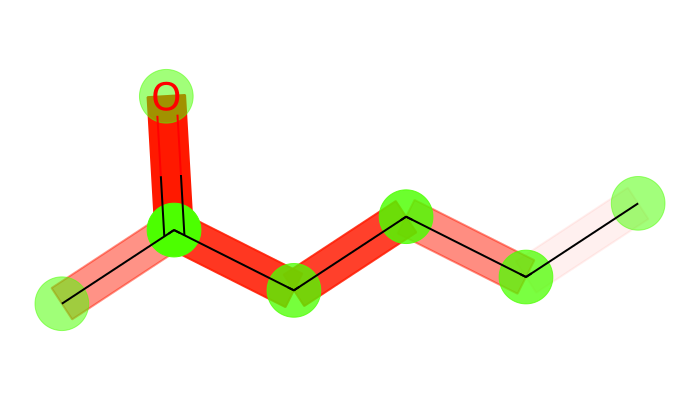}
        \end{minipage}
        \hfill
        \begin{minipage}[b]{0.24\textwidth}
            \includegraphics[width=\textwidth]{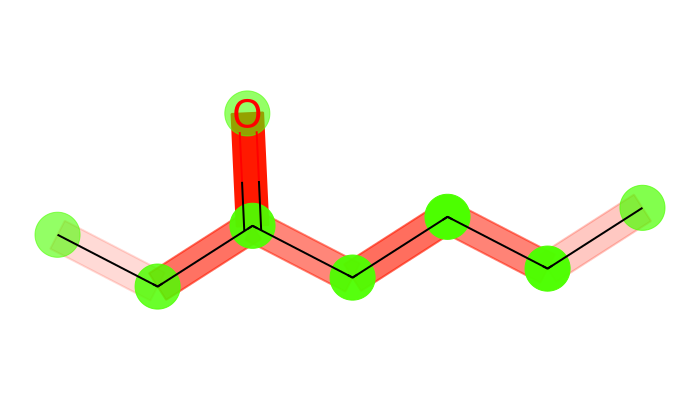}
        \end{minipage}
        \hfill
        \begin{minipage}[b]{0.24\textwidth}
            \includegraphics[width=\textwidth]{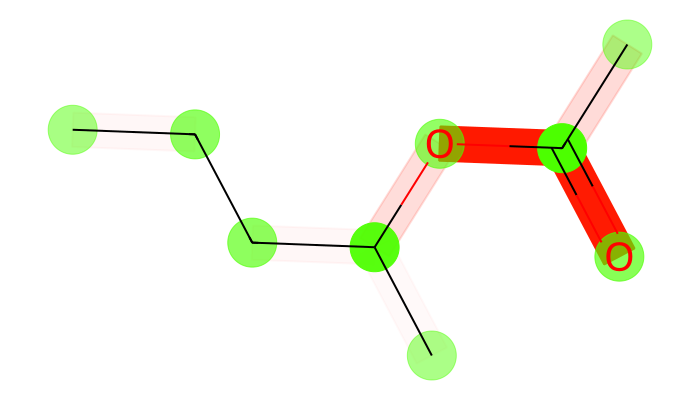}
        \end{minipage}
        \hfill
        \begin{minipage}[b]{0.24\textwidth}
            \includegraphics[width=\textwidth]{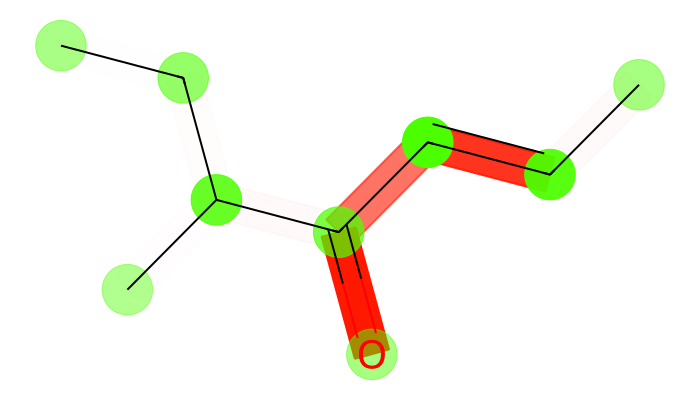}
        \end{minipage}
        \caption{}
        \label{fig:ketonic_row}
    \end{subfigure}
\caption{\textbf{Figure S4: Explainability through integrated gradients\cite{sundararajan2017axiomatic} for different odor descriptors:} (a) alcohol, (b) sulfurous, (c) phenolic, (d) fruity, (e) musk, and (f) ketonic. Green indicates atom-level contributions; red, bond-level contributions. Opacity indicates node and edge importance and is proportional to the values.}
    \label{fig:odor_explainability}
\end{figure}
\bibliography{achemso-demo}
\end{document}